\documentclass[letterpaper, 10 pt, conference]{ieeeconf}  

\IEEEoverridecommandlockouts                              

\usepackage{graphics} 
\usepackage{subfigure}
\usepackage{bm}
\usepackage{float} 
\usepackage{stfloats}
\usepackage{caption}
\usepackage{cancel}
\usepackage{widetext}
\usepackage{epsfig} 
\usepackage{overpic}
\usepackage{times} 
\usepackage{booktabs}
\usepackage{rotating}
\usepackage{algorithmic}
\usepackage{algorithm}

\usepackage{amsmath} 
\usepackage{amssymb}  

\usepackage{amsthm}
\usepackage{mathrsfs}
\usepackage[table, dvipsnames, svgnames, x11names]{xcolor} 
\usepackage{colortbl}
\usepackage{multirow}
\usepackage{marvosym}
\definecolor{mygray}{gray}{.9}
\usepackage{pdflscape}
\usepackage{pifont}
\usepackage{cite}
\let\labelindent\relax
\usepackage{enumitem}
\usepackage[colorlinks,
            linkcolor=blue,
            anchorcolor=blue,
            citecolor=green,
            urlcolor=violet,
            backref=page]{hyperref}
\title{\LARGE \bf
Continual Driving Policy Optimization with\\Closed-Loop Individualized Curricula}

\author{Haoyi Niu$^{1\dag}$, Yizhou Xu$^{1\dag}$, Xingjian Jiang$^{1}$, Jianming Hu$^{1}$\textsuperscript{\Letter}
\thanks{\dag Work done with equal contribution. Email: 
{\tt\small \{nhy22, xyz20\}@mails.tsinghua.edu.cn}}
\thanks{$^{1}$Department of Automation, Tsinghua University, Beijing 100084, China. Correspondence to: 
Jianming Hu. {\tt\small hujm@mail.tsinghua.edu.cn}}
\thanks{Source code and supplementary materials are available at \url{https://sites.google.com/view/icra2024clic}.}
}

\begin{document}
\maketitle
\thispagestyle{empty}
\pagestyle{empty}

\begin{abstract}
    The safety of autonomous vehicles (AV) has been a long-standing top concern, stemming from the absence of rare and safety-critical scenarios in the long-tail naturalistic driving distribution.
    To tackle this challenge, a surge of research in scenario-based autonomous driving has emerged, with a focus on generating high-risk driving scenarios and applying them to conduct safety-critical testing of AV models.
    However, limited work has been explored on the reuse of these extensive scenarios to iteratively improve AV models.
    Moreover, it remains intractable and challenging to filter through gigantic scenario libraries collected from other AV models with distinct behaviors, attempting to extract transferable information for current AV improvement.
    Therefore, we develop a continual driving policy optimization framework featuring \underline{C}losed-\underline{L}oop \underline{I}ndividualized \underline{C}urricula (CLIC), which we factorize into a set of standardized sub-modules for flexible implementation choices: \textit{AV Evaluation}, \textit{Scenario Selection}, and \textit{AV Training}.
    CLIC frames AV Evaluation as a collision prediction task, where it estimates the chance of AV failures in these scenarios at each iteration.
    Subsequently, by re-sampling from historical scenarios based on these failure probabilities, CLIC tailors individualized curricula for downstream training, aligning them with the evaluated capability of AV.
    Accordingly, CLIC not only maximizes the utilization of the vast pre-collected scenario library for closed-loop driving policy optimization but also facilitates AV improvement by individualizing its training with more challenging cases out of those poorly organized scenarios.
    Experimental results clearly indicate that CLIC surpasses other curriculum-based training strategies, showing substantial improvement in managing risky scenarios, while still maintaining proficiency in handling simpler cases.
\end{abstract}

\vspace{2.5mm}
\section{Introduction}

With remarkable advancements in deep learning (DL) and deep reinforcement learning (DRL), autonomous driving has gained substantial interest from academia, industry, and the public.
However, the deployment of autonomous vehicles (AV) in the real world has been significantly impeded by safety concerns. 
The primary crux sources from the distribution of naturalistic driving data (NDD), which exhibits a long-tailed pattern~\cite{zhang2023deep}, leading to a severe data imbalance characterized by a scarcity of safety-critical scenarios.
Naturally, solely relying on NDD would require training and testing AV for billions of miles~\cite{kalra2016driving} to ensure safety.
Thereby, a repertoire of studies has emphasized the need to generate safety-critical scenarios~\cite{wang2021advsim,ding2020learning,feng2020testing1,feng2020testing2,feng2020testing3,sun2021corner,niu2021dr2l,lee2020adaptive,rempe2022generating,niu2023re} to address the data imbalance issue.
This has led to the emergence of an exciting avenue coined as scenario-based autonomous driving~\cite{nalic2020scenario, li2020scenario} that harbors several advantages against autonomous driving under naturalistic situations.
Logged scenarios offer substantial reproducibility, controllability and flexibility for re-organization, avoiding repetitive computations during replay and allowing re-sampling of valuable scenarios at will.
These advantageous properties hold great promise for accelerating the testing phase~\cite{feng2023dense, zhao2016accelerated, zhao2017accelerated, huang2017accelerated,zhong2021survey, fremont2020formal,yang2023adaptive} introduced with advanced re-sampling techniques over safety-critical scenarios.



However, limited research has explored the potential of utilizing the rich and diverse historical scenarios for closed-loop training of AV, rather than just for testing purposes.
The main challenge is twofold: (1) \textit{Scenario Transferability}: Pre-collected scenarios can vary significantly across different driving patterns, including those generated by overly timid and considerably aggressive AV models or human drivers. AV models trained directly on such scenarios may not always yield improvement due to the significant distribution inconsistency~\cite{cao2021confidence, ding2023survey}.
(2) \textit{Scenario Adaptability}: Unlike the fixed AV model used during the testing phase, closed-loop AV policy optimization inevitably involves dynamical improvement of the iterated AV model. Moreover, driving scenarios are widely acknowledged for their diversity in difficulty levels, so the AV model needs to be fed with carefully selected scenarios that align with their capabilities at each iteration~\cite{ding2023survey}.
This highlights the need to offer individualized curricula comprising more challenging scenarios and less boring ones, tailored to fulfill the requirements of efficient and effective training at current stage.

\begin{figure*}[t]
    \centering
    \begin{overpic}[width=1\linewidth]{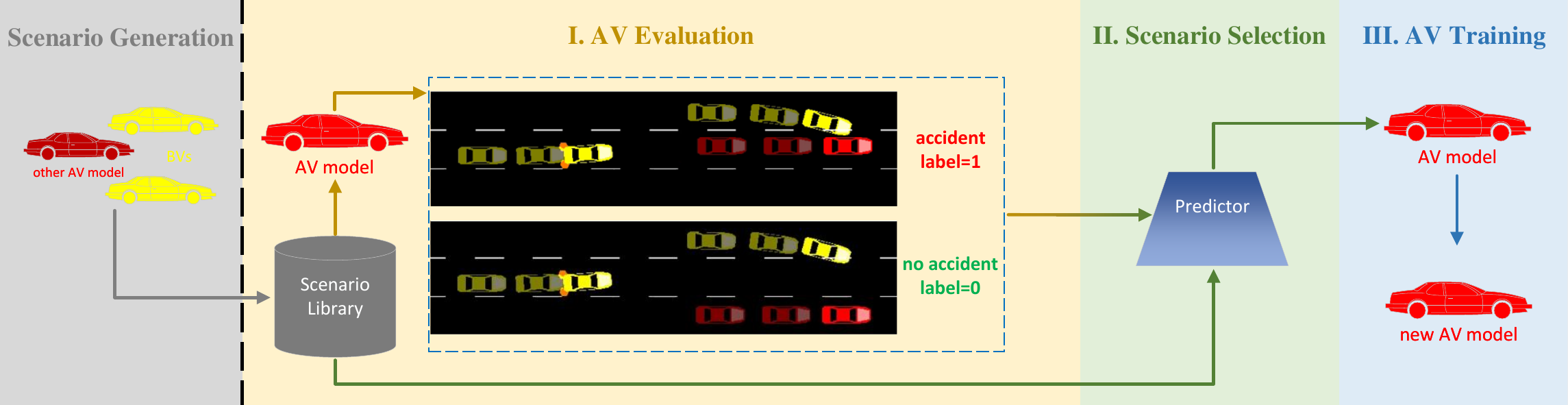}
        \put(1,18.5){\small\bfseries{$\mu(\mathbf{a}|\mathbf{s})$}}
        \put(9.5,8){\small\bfseries{$\{d^{(j)}\}$}}
        \put(17,21){\small\bfseries{$\pi_{\phi}(\mathbf{a}|\mathbf{s})$}}
        \put(22,12){\small\bfseries{$d_{\rm eval}$}}
        \put(20.6,3.75){\small\bfseries{$\mathcal{D}$}}
        \put(65,13.25){\small\bfseries{$d_{\rm eval}$}}
        \put(65.75,10.25){\small\bfseries{$l_{\rm gt}$}}
        \put(74.5,10){\small\bfseries{$p_{\psi}(\cdot|\phi)$}}
        \put(80,19){\small\bfseries{$d_{\rm train}$}}
        \put(90,8.75){\small\bfseries{$\pi_{\phi^{\prime}}(\mathbf{a}|\mathbf{s})$}}
        \put(90,20){\small\bfseries{$\pi_{\phi}(\mathbf{a}|\mathbf{s})$}}
    \end{overpic}
    \caption{Overview of CLIC Algorithmic Architecture}
    \label{fig:algorithm_flow}
    \vspace{-5mm}
\end{figure*}
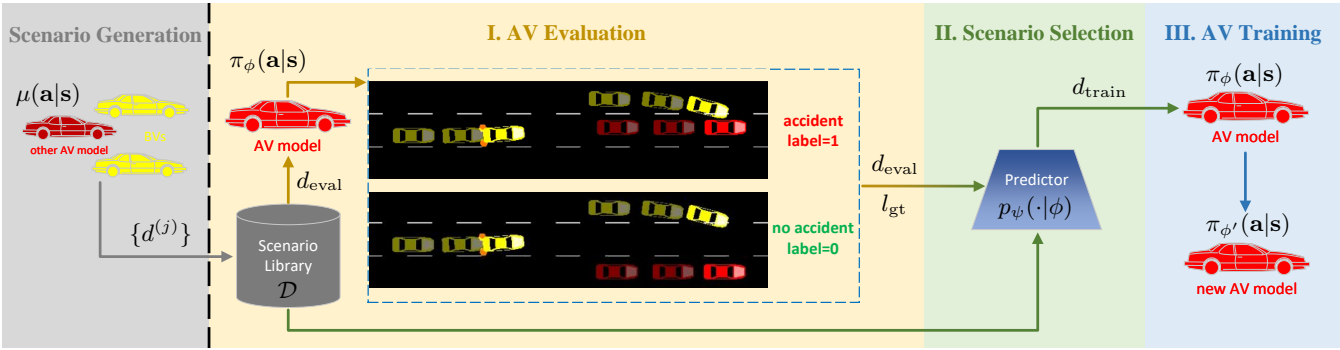

To tackle these challenges, we introduce a novel framework of \textbf{Continual Driving Policy Optimization with \underline{C}losed-\underline{L}oop \underline{I}ndividualized \underline{C}urricula (CLIC)}, which we divide into three standardized sub-modules for flexible implementation choices, as illustrated in Figure~\ref{fig:algorithm_flow}: 
(1) \textbf{AV Evaluation}: At each iteration, 
we begin by exposing the AV model to a subset of scenarios to assess the current AV capability. 
(2) \textbf{Scenario Selection}: Next, we aim to estimate whether the AV model has an accident in each scenario. To achieve this, we employ a discriminator network trained to predict accident probabilities based on the assessed outcomes, which we term ``difficulty predictor". Subsequently, we leverage the predicted labels indicating risk levels to reweight sampling within the scenario library. This provides AV with individualized curricula that align with its current capability.
(3) \textbf{AV Training}: The AV model is then trained using the scenarios obtained through individualized re-sampling.
Overall, CLIC adheres to the principles of continual and curriculum learning by commencing training with simpler samples and progressively introducing more challenging ones, achieving scenario transfer and adaptation out of those diverse yet poorly organized historical scenario libraries for closed-loop AV optimization. The algorithmic design also helps safeguard against catastrophic forgetting~\cite{toneva2018empirical,khetarpal2022towards} of AV model.
Against several competing curriculum-based baselines, experimental results demonstrate that CLIC effectively optimizes AV models to handle more challenging scenarios while minimizing any degradation in performance for easier cases.

\vspace{1.2mm}
\section{Related Work}\label{2}
\vspace{1mm}
\subsection{Scenario-Based Autonomous Driving}\label{2-2}
Scenario-based autonomous driving has recently become a popular paradigm for training and testing autonomous driving models. Currently, the focus of research in this area mainly lies in scenario generation and tesing AV model, particularly generating extreme scenarios which are rare in NDD by training RL models to control background vehicle (BV). Adaptive Stress Testing~\cite{koren2018adaptive, koren2019efficient, corso2019adaptive, lee2020adaptive} adopts Monte Carlo tree search and DRL to generate extreme scenarios. Feng et al.~\cite{feng2020testing1, feng2020testing2, feng2020testing3, sun2021corner} propose a unified framework for adaptive testing scenario library generation. Bayesian optimization and DRL are employed in different cases. AdvSim~\cite{wang2021advsim} and KING~\cite{hanselmann2022king} use a constructed adversarial cost function to explore and generate safety-critical scenarios. (Re)$^2$H2O~\cite{niu2023re} efficiently generates varied adversarial scenarios by combining NDD with simulation data through hybrid offline-and-online RL. 

The scenarios generated from the above methods can be stored by saving the state information of all vehicles at each time step as a static scenario library, which is stable and easy to reproduce. However, according to our research, there is currently a lack of work that utilizes existing large-scale static scenario libraries for training AV models, thus failing to establish an industrial closed-loop of scenario generation, AV training and testing. Our goal is to complete the missing step in this closed loop, achieving an integrated approach to train and test scenario-based AV models.

\subsection{Curriculum Learning}\label{2-1}
Curriculum learning (CL)~\cite{bengio2009curriculum} is a training strategy involving reweighting the training data and designing a series of tasks or examples in increasing difficulty order. By gradually exposing the model to more challenging instances, CL enables a smoother learning process and better generalization performance, prevent the model from becoming trapped in local optima. In addition to its original definition, CL has been extended to a series of similar or expanded algorithms, such as self-paced learning (SPL)~\cite{kumar2010self, jiang2015self, ren2018self, klink2020self} and teacher-guided learning~\cite{matiisen2019teacher, portelas2020teacher, romac2021teachmyagent, schraner2022teacher, shenfeld2023tgrl}. These approaches have found wide applications in both DL~\cite{graves2017automated, jiang2018mentornet, hacohen2019power, kim2018screenernet} and RL~\cite{florensa2017reverse, florensa2018automatic, klink2021boosted, cai2023curriculum} domains. In addition, Prioritized Experience Replay (PER)~\cite{schaul2015prioritized} is another method in RL that involves data reweighting. Instead of reweighting each training example, it focuses on reweighting individual transitions, assigning higher priority to transitions that are deemed more important for learning.

The application of CL in AV training is currently not very common, and most of them align with predefined CL~\cite{wang2021survey}, and distinguish curricula by discrete factors such as weather and road topology structure~\cite{ozturk2021investigating}, providing future BV trajectories of different lengths~\cite{khaitan2022state}, changing the target driving distance~\cite{agarwal2021sparse} and the quantity of BV~\cite{anzalone2021reinforced, anzalone2022end}. 
In addition,~\cite{qiao2018automatically} utilize the value function $V$ from RL to measure the learning potential of different tasks, and divide tasks based on the distance range from the starting point to the intersection. Most of the aforementioned works define the curriculum as discrete stages and predefine when to switch between them, which often fall short in terms of automaticity, adaptability and flexibility; and usually train the model with random traffic flow. Although~\cite{khaitan2022state} also employs scenario-based training, it is not applicable to our large-scale scenario library with varying levels of difficulty. 

To overcome the aforementioned limitations and better adapt to our objectives and the existing scenario library, we employ the definition of \textbf{Data-level Generalized CL}~\cite{wang2021survey}: Curriculum learning refers to the reweighting of the target training set distribution in $T$ training steps. We aim to assign different weights to each scenario in the scenario library automatically, and then perform weighted sampling to obtain training scenarios as individualized curriculum. 
By continually updating these weights, we can achieve changes in the difficulty distribution of the static scenario library, thus enabling continual optimization of driving policy. 

\section{Methodology}\label{3}
\subsection{Problem Formulation}\label{3-1}
\subsubsection{Scenario Formulation}\label{3-1-1}
Our scenarios are based on driving straight on the highway and assume that within our area of interest, there is only one AV ($V^{(0)}$) and $N$ BVs ($V^{(1)},\ldots,V^{(N)}$). For each vehicle at moment $t$, we use its lateral and longitudinal position $(x,y)$, velocity $v$ and heading angle $\theta$ as the state vector, and the changes in AV's velocity and heading angle between two moments as the action vector:
    $\mathbf{s}^{(i)}_t=[x^{(i)}_t,y^{(i)}_t,v^{(i)}_t,\theta^{(i)}_t],\quad \mathbf{a}^{(0)}_t=[\Delta v^{(0)}_t,\Delta\theta^{(0)}_t]$. 
In this way, the complete state and action vectors are $\mathbf{s}_t=[\mathbf{s}^{(0)}_t,\mathbf{s}^{(1)}_t,\ldots,\mathbf{s}^{(N)}_t]^{\rm T},\mathbf{a}_t=[\mathbf{a}^{(0)}_t]^{\rm T}$, 
while $\mathbf{s}_t^{\rm BV}=[\mathbf{s}^{(1)}_t,\ldots,\mathbf{s}^{(N)}_t]^{\rm T}$ only represents the state of all BVs.

A scenario $d$ is defined as a finite sequence of traffic scenes consisting of successive $H$ frames, where each frame contains the state of all BVs, and the initial state of all traffic participants (including the AV and all BVs) is given at the initial time: 
\begin{equation}\label{scenario_definition}
    d=[\mathbf{s}_0,\mathbf{s}_1^{\rm BV},\ldots,\mathbf{s}_H^{\rm BV}]
\end{equation}
It should be emphasized that this definition of scenarios is AV-agnostic, getting rid of the influence of other AV models with distinct behaviors used to collect those scenarios. 
Therefore, they can be reapplied to different AV models for training. 

Due to the limited precision of the decision modeling at the trajectory level in the SUMO simulator~\cite{behrisch2011sumo} we use, we choose to manually calculate the kinematic state transitions of AV:
\begin{equation}
    \begin{aligned}
        x^{(0)}_{t+\Delta t}&=x^{(0)}_t+v^{(0)}_t\cdot \cos\theta^{(0)}_t\cdot\Delta t\\
        y^{(0)}_{t+\Delta t}&=y^{(0)}_t+v^{(0)}_t\cdot \sin\theta^{(0)}_t\cdot\Delta t\\
        v^{(0)}_{t+\Delta t}&=v^{(0)}_t+\Delta v^{(0)}_t\\
        \theta^{(0)}_{t+\Delta t}&=\theta^{(0)}_t+\Delta \theta^{(0)}_t
    \end{aligned}
\end{equation}
while the state transition of BVs is directly provided by the recorded scenario data in Equation~\ref{scenario_definition}. 

\subsubsection{Training AV with RL}\label{3-1-2}
Consider the decision of AV in a traffic environment as a Markov Decision Process (MDP) defined by a tuple $(\mathcal{S},\mathcal{A},\mathcal{R},P,\rho,\gamma)$~\cite{sutton1998introduction, puterman2014markov}, where $\mathcal{S}$, $\mathcal{A}$ and $P$ are the state space, action space and transition probability as outlined in Section~\ref{3-1-1}, and $\rho$ and $\gamma$ are the initial state distribution and discount factor. Reward function $r_t=r(\mathbf{s}_t,\mathbf{a}_t)\in\mathcal{R}$ is defined as 
\begin{equation}\label{reward}
    r_t=r_{acc}+r_{vel}+r_{yaw}+r_{lane}
\end{equation}
which includes the following items (the coefficient values for each item can be found in Table~\ref{table_reward_coefficient}): 

\begin{itemize}[leftmargin=*,topsep=0pt]
\item\textit{Accident}: Instruct AV to avoid accidents, including collisions with BV and driving off roads: 
\begin{equation}\label{reward_acc}
    r_{acc}=-\rho_{acc}\cdot\mathbb{I}_{\rm \{AV\in\mathcal{C}\}}
\end{equation}
where $\mathbb{I}$ represents the indicative function, $\mathcal{C}$ is the set of vehicles that have an accident within the current time step, and $\rho_{acc}$ is the corresponding coefficient.

\item\textit{Velocity}: Encourage AV to drive faster within the allowed velocity range $[v_{min}, v_{max}]$: 
\begin{equation}\label{reward_vel}
    r_{vel}=\rho_{vel}\cdot\frac{v_t^{(0)}-(v_{max}+v_{min})/2}{(v_{max}-v_{min})/2}
\end{equation}
where $v_t^{(0)}$ is the current velocity of AV, and $\rho_{vel}$ is the corresponding coefficient.

\item\textit{Heading direction}: Instruct AV to drive smoothly and drive along the road direction: 
\begin{equation}\label{reward_yaw}
    r_{yaw}=-\rho_{yaw}\cdot\lvert\theta_t^{(0)}\rvert
\end{equation}
where $\theta_t^{(0)}$ is the current heading angle of AV, and $\rho_{yaw}$ is the corresponding coefficient.

\item\textit{Lane selection}: Encourage AV to drive on the ``best'' lane, where the distance between AV and the nearest BV in front of AV on the same lane is the maximum: 
\begin{equation}\label{reward_lane}
    r_{lane}=\rho_{lane}\cdot\mathbb{I}_{\rm \{AV\ on\ the\ best\ lane\}}
\end{equation}
where $\rho_{lane}$ is the corresponding coefficient.
\end{itemize}

Although our scenario library is static, the transition distribution used for training varies as the AV policy changes, because the AV state information is not included in the scenario library but is instead populated through online rollouts to align with the current AV policy. Thus, we choose the online RL algorithm Soft Actor-Critic (SAC)~\cite{haarnoja2018softarxiv, haarnoja2018softicml} to solve the MDP problem. The objective function of SAC includes an entropy term $\mathcal{H}$ of the AV policy distribution: 
\begin{equation}
\begin{aligned}
    J_{\pi}(\phi)=\sum_{t=0}^H\ &\mathbb{E}_{\mathbf{s}_0\sim\rho,\mathbf{a}_t\sim\pi_{\phi}(\cdot|\mathbf{s}_t),\mathbf{s}_{t+1}\sim P(\cdot|\mathbf{s}_t,\mathbf{a}_t)} \\
    &\left[\gamma^tr(\mathbf{s}_t,\mathbf{a}_t)+\alpha\mathcal{H}(\pi_{\phi}(\cdot|\mathbf{s}_t))\right]
\end{aligned}
\end{equation}
where $\pi_{\phi}$ is the AV policy parameterized by $\phi$, 
and $\alpha$ is the temperature hyperparameter. 
This encourages the exploration of more possible actions while maximizing the expected reward discounted by $\gamma$, which enhances the robustness of the model and is also more consistent with the application patterns in real-world traffic scenarios.

\begin{algorithm}[t]
\caption{Universal Algorithmic Framework}
\label{meta_algorithm}
    \begin{algorithmic}[1]
        \STATE \textbf{Initialize}: AV policy $\pi_{\phi}(\mathbf{a}|\mathbf{s})$
        \STATE \textbf{Input}: static scenario library $\mathcal{D}$
        \FOR{each iteration}
            \STATE $\texttt{info}\leftarrow\textsc{EvaluateAV}(\pi_{\phi}(\mathbf{a}|\mathbf{s}),\mathcal{D})$
            \STATE $d_{\rm train} \leftarrow \textsc{SelectScenario}(\mathcal{D},\texttt{info})$
            \STATE $\pi_{\phi}(\mathbf{a}|\mathbf{s})\leftarrow\textsc{TrainAV}(\pi_{\phi}(\mathbf{a}|\mathbf{s}),d_{\rm train})$
        \ENDFOR
    \end{algorithmic}
\end{algorithm}

\subsection{Algorithmic Framework}\label{3-2}
We refine our approach into a universal modular framework (Algorithm~\ref{meta_algorithm}), and provide our specific definitions for each module based on this in Section~\ref{3-3}.

In our universal algorithmic framework, each module is undefined and expandable, which allows us to focus on the relationships between modules and the flow of data, while ignoring the implementation details of each module. It also facilitates the adoption of different specific implementations within our framework in order to propose new methods.

Our universal algorithmic framework consists of three main stages: AV Evaluation, Scenario Selection, and AV Training. In the \textbf{AV Evaluation} stage, we evaluate the current AV model to obtain an estimate of the current capabilities of the AV model, denoted as \texttt{info} in Algorithm~\ref{meta_algorithm}. The \texttt{info} here can take any form of evaluation information, depending on how the evaluation process and its corresponding output are defined. In the \textbf{Scenario Selection} stage, training scenarios are selected from the scenario library based on a certain criterion, in order to explore the boundaries of the AV model's capabilities and achieve the maximum improvement in AV performance. In the \textbf{AV Training} stage, any RL algorithm can be used to improve the AV policy in the training scenarios. These three stages are executed sequentially and stop after several iterations.

\subsection{Algorithmic Implementation}\label{3-3}
\subsubsection{AV Evaluation}\label{3-3-1}
In this stage, we aim to evaluate the capabilities of the current AV model $\pi_{\phi}(\mathbf{a}|\mathbf{s})$ using a subset of scenarios $d_{\rm eval}$ from the scenario library. Considering that the time cost of this stage should not be too high and the unbiasedness of the scenario distribution should be maintained, we randomly sample $m$ scenarios from the scenario library $\mathcal{D}$ with a total of $M$ scenarios, and sequentially rollout them in the environment to interact with the current AV model. Then we return the results of the interaction, which are a set of labels $l_{\rm gt}$ indicating whether an accident has occurred in AV. Detailed algorithmic steps are shown in Algorithm~\ref{evaluate_av}. 

\begin{algorithm}[t]
\caption{\textsc{EvaluateAV}}
\label{evaluate_av}
    \begin{algorithmic}[1]
        \STATE \textbf{Initialize}: evaluation scenario number $m$
        \STATE \textbf{Input}: current AV policy $\pi_{\phi}(\mathbf{a}|\mathbf{s})$, scenario library $\mathcal{D}$
        \STATE $d_{\rm eval}\leftarrow \textsc{Sample}(\mathcal{D},m)$
        \FOR{$d^{(j)}$ in $d_{\rm eval}$}
            \STATE $\mathbf{s}\leftarrow d_{\mathbf{s}_0}^{(j)};\ \texttt{done}\leftarrow {\rm False}$
            \WHILE{not \texttt{done}}
                \STATE $\mathbf{a}\leftarrow \arg\max_\mathbf{a} \pi_{\phi}(\mathbf{a}|\mathbf{s})$
                \STATE $\mathbf{s}^{\prime},r,\texttt{done}\leftarrow \textsc{Step}(\mathbf{a})$;\ $\mathbf{s}\leftarrow \mathbf{s}^{\prime}$
            \ENDWHILE
            \STATE $l_{\rm gt}^{(j)}\leftarrow \mathbb{I}_{\{{\rm AV\in\mathcal{C}\}}}$
        \ENDFOR
        \RETURN $l_{\rm gt}$
    \end{algorithmic}
\end{algorithm}

\subsubsection{Scenario Selection}\label{3-3-2}
In order to continually optimize driving policy and enhance the ability of AV to handle extreme scenarios, it is important to focus training on more challenging scenarios while including a small number of less difficult scenarios to prevent forgetting. Thus, associating the difficulty of each scenario with its sampling weight is an intuitive approach. 
However, currently there is no perfect benchmark to assess scenario difficulty, meanwhile scenario difficulty is also a rather individualized and non-universal metric, which is related to the AV itself. Thus we make use of the results from each evaluation stage as the information for the current AV model. We treat the scenarios used for evaluation $d_{\rm eval}$ and their collision labels $l_{\rm gt}$ as the ground truth, and then employ a supervised learning approach to train a difficulty predictor model $p_{\psi}(d|\phi)$, where $d$ can be any scenario that meets the definition in Equation~\ref{scenario_definition}. 
We utilize a Multi-Layer Perceptron (MLP)~\cite{rumelhart1986learning} parameterized by $\psi$ as the difficulty predictor $p_{\psi}(\cdot|\phi)$\footnote{Refer to Appendix~\ref{appendix_predictor} for comparison of different predictor models.}, and flatten the entire scenario data $d$ into a one-dimensional vector as the input of the predictor. Binary Cross-Entropy Loss (BCE Loss) is used as the loss function: 
\begin{equation}\label{bceloss}
\small
    \mathcal{L}_{\rm pred}(\psi)=-\mathbb{E}_{d^{(j)}\sim d_{\rm eval}}\big[l_{\rm gt}^{(j)}\cdot\log(l_{\rm pred}^{(j)})+(1-l_{\rm gt}^{(j)})\cdot\log(1-l_{\rm pred}^{(j)})\big]
\end{equation}
where $l_{\rm pred}^{(j)}=p_{\psi}(d^{(j)}|\phi)$ is the predicted label value of the $j$-th scenario $d^{(j)}$. The number of training epochs is determined through experiments to ensure that the predictor neither underfits nor overfits the training data. The predictor is then applied to the entire scenario library $\mathcal{D}$ to obtain the predicted label values $l_{\rm all}$ for each scenario. These predicted label values are then used as weights for weighted sampling, resulting in a batch of training scenarios $d_{\rm train}$. Specifically, the probability of sampling the $j$-th scenario $d^{(j)}$ is: 
\begin{equation}\label{weighted_sample}
    P(d^{(j)}\in d_{\rm train})=\frac{l_{\rm all}^{(j)}}{\sum_{i=1}^M l_{\rm all}^{(j)}}=\frac{p_{\psi}(d^{(j)}|\phi)}{\sum_{i=1}^M p_{\psi}(d^{(j)}|\phi)}
\end{equation}
Algorithm~\ref{evaluate_av} and Algorithm~\ref{select_scenario} together form the process of individualized curriculum design. 

\begin{algorithm}[t]
\caption{\textsc{SelectScenario}}
\label{select_scenario}
    \begin{algorithmic}[1]
        \STATE \textbf{Initialize}: difficulty predictor $p_{\psi}(\cdot|\phi)$, learning rate $\lambda$, training scenario number $n$
        \STATE \textbf{Input}: scenario library $\mathcal{D}$, evaluation scenarios $d_{\rm eval}$, labels of evaluation scenarios $l_{\rm gt}$
        \FOR{each epoch}
            \STATE $l_{\rm pred}\leftarrow p_{\psi}(d_{\rm eval}|\phi)$
            \STATE $\psi \leftarrow \psi - \lambda\nabla_{\psi}\mathcal{L}_{\rm pred}(\psi)$ \COMMENT{Equation~\ref{bceloss}}
        \ENDFOR
        \STATE $l_{\rm all}\leftarrow p_{\psi}(\mathcal{D}|\phi)$
        \STATE $d_{\rm train}\leftarrow \textsc{WeightedSample}(\mathcal{D},n,l_{\rm all})$ \COMMENT{Equation~\ref{weighted_sample}}
        \RETURN $d_{\rm train}$
    \end{algorithmic}
\end{algorithm}

\subsubsection{AV Training}\label{3-3-3}
In this stage, we employ online RL algorithm SAC to train the AV model. At each epoch, we first shuffle the order of the training scenarios and then rollout each scenario one by one, updating the parameters of the AV model based on SAC, as described in Section~\ref{3-1-2}. 


By incorporating the specific implementation of the aforementioned module into the universal algorithmic framework, we obtain an overall architecture illustrated in Figure~\ref{fig:algorithm_flow}, which includes the three modules mentioned above. 
The scenario generation stage for other AV model $\mu(\mathbf{a}|\mathbf{s})$ on the left of the figure has already been extensively studied and is therefore outside the scope of this algorithm. In each training iteration, a batch of scenarios $d_{\rm eval}$ is randomly sampled from the static scenario library $\mathcal{D}$ to evaluate the current AV model. These scenarios are rolled out in the environment and interacts with AV to obtain accident labels $l_{\rm gt}$. These scenarios and labels are then used to train the difficulty predictor $p_{\psi}(\cdot|\phi)$, which, after training, predicts labels for all scenarios in $\mathcal{D}$. These predicted labels $l_{\rm all}$ are used as weights to perform weighted sampling on all scenarios, resulting in the training scenarios $d_{\rm train}$. Finally, the AV model $\pi_{\phi}(\mathbf{a}|\mathbf{s})$ is trained using an online RL algorithm on these training scenarios, completing a full closed-loop training iteration. 

\begin{figure*}[t]
    \centering
    \includegraphics[width=1\linewidth]{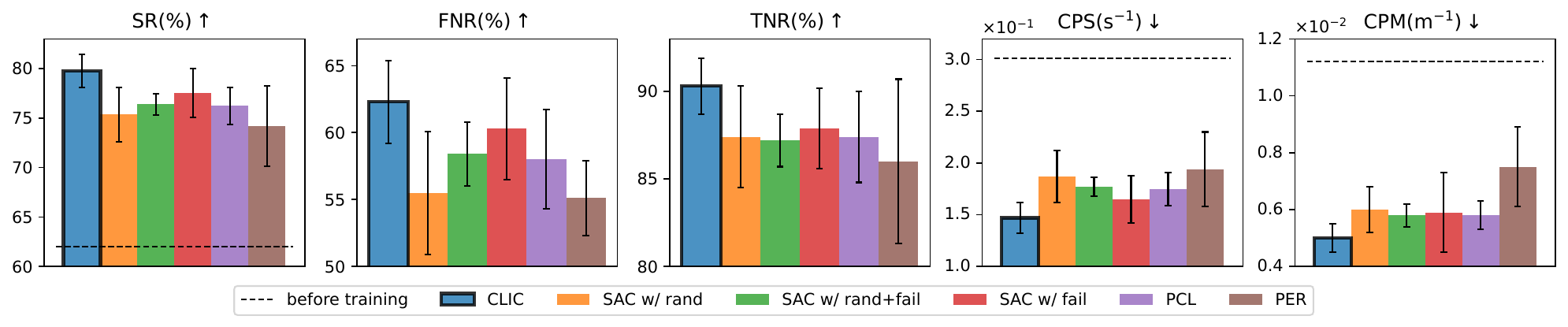}
    \caption{Comparison of various metrics for CLIC and baselines.}
    \label{fig:metrics}
\end{figure*}

\section{Experiments}\label{4}
We provide evidence for the superiority of CLIC in training AV models by addressing the following questions: 
\begin{itemize}[leftmargin=*,topsep=0pt]
    \item Can CLIC train safer AV models than other baselines? (Section~\ref{4-1})
    \item Is CLIC strengthening the AV model while increasing the difficulty of training scenarios selected by the predictor throughout the training process? (Section~\ref{4-2})
    \item How does CLIC reweight the data distribution of the whole scenario library according to the current AV capability? (Section~\ref{4-3})
    \item Can CLIC select training scenarios specifically tailored to the defects of a particular vehicle? (Section~\ref{4-4})
\end{itemize}

\subsection{Comparison Experiment}\label{4-1}
\subsubsection{Experimental Settings}\label{4-1-1}
The comparative experiment is our main experiment. Here we first explain the experimental settings such as dataset, baselines and metrics. 
\paragraph{Dataset}\label{4-1-1-1}
All traffic scenarios used in our experiments are generated by (Re)$^2$H2O~\cite{niu2023re}. Refer to Appendix~\ref{appendix_dataset} for more details about the dataset.

\paragraph{Baselines}\label{4-1-1-2}
We select five common curriculum related baselines for comparison: 
\begin{itemize}[leftmargin=*,topsep=0pt]
    \item \textbf{\textit{SAC w/ rand}}: Simply sample a batch of training scenarios randomly from the scenario library.
    \item \textbf{\textit{SAC w/ rand+fail}}: In each training iteration, half of the training scenarios are randomly sampled from the scenario library, while another half consists of scenarios that failed in the corresponding AV Evaluation stage.
    \item \textbf{\textit{SAC w/ fail}}: All training scenarios are chosen from those that fail in AV Evaluation stage.
    \item \textbf{\textit{PCL}}: Following the idea of predefined curriculum learning (PCL)~\cite{wang2021survey}, we divide the training process into four phases of the curriculum based on the number of BV (1$\sim$4). The training set for each phase adds scenarios with a higher number of BV based on the previous phase.
    \item \textbf{\textit{PER}}: Apply PER~\cite{schaul2015prioritized} into RL based on \textit{SAC w/ rand}.
\end{itemize}

\paragraph{Evaluation Metrics}\label{4-1-1-3}
To evaluate the capabilities and training effectiveness of the AV model in extreme scenarios, we employed the following metrics\footnote{We expect higher values for SR, FNR and TNR, and lower values for CPS and CPM, which indicates that the AV model is safer.}:  
\begin{itemize}[leftmargin=*,topsep=0pt]
    \item \textbf{SR} (\%): Success rate, referring to the success rate of the AV model in all scenarios after the final iteration of training. It reflects the overall safety performance of the AV model in various extreme scenarios.
    \item \textbf{Confusion Matrix}: To distinguish between the difficulty levels of different scenarios and to highlight the effectiveness of the AV model in high-difficulty scenarios, we also borrow the concept of confusion matrix in supervised learning: we define the label of successful scenarios as 0, and that of failed scenarios as 1. The test results before training are treated as ground truth, while those after training are treated as predicted values. Specifically, we compute the following two metrics:
    \begin{itemize}
        \item  \textbf{FNR} (\%) $=\frac{{\rm FN}}{{\rm TP}+{\rm FN}}$: the proportion of successful scenarios after training that are failed before.
        \item \textbf{TNR} (\%) $=\frac{{\rm TN}}{{\rm TN}+{\rm FP}}$: the proportion of successful scenarios after training that are successful before.
    \end{itemize}
    \item \textbf{CPS} (${\rm s}^{-1}$) and \textbf{CPM} (${\rm m}^{-1}$): As defined in~\cite{niu2023re}, CPS is Average Collision Frequency Per Second and CPM is Average Collision Frequency Per Meter, calculated as following: 
    \begin{equation}
        {\rm CPS}=\frac{N_{acc}}{T_{total}},\quad {\rm CPM}=\frac{N_{acc}}{D_{total}}
    \end{equation}
    where $N_{acc}$ is the number of scenarios in which AV has an accident during testing, $T_{total}$ and $D_{total}$ are the total testing time and the total driving distance traveled by AV along the road direction.
\end{itemize}


\subsubsection{Experimental Results}\label{4-1-2}
We conduct tests on all scenarios in the scenario library to compare different baselines on the aforementioned metrics, and present the results\footnote{The complete results are shown in Table~\ref{table_complete_results_safety}.} from 5 random seeds as shown in Figure~\ref{fig:metrics}. CLIC outperforms all baselines in all metrics, achieving the highest SR, FNR, TNR and the lowest CPS and CPM. This indicates that our training method not only enables the AV model to learn more challenging scenarios it couldn't handle before but also minimizes the risk of forgetting previously learned scenarios to the utmost. Although \textit{SAC w/ fail} focuses on training failure scenarios and achieves relatively good results on FNR, it is also easy to forget simple scenarios, leading to a little bad performance on TNR. Furthermore, \textit{SAC w/ rand} and \textit{PER} perform the worst, further illustrating the importance of a well-designed curriculum compared to blindly training on randomly sampled scenarios from the scenario library. We also included the SR, CPS, and CPM of the AV model with random initialization before training in the figure. It can be observed that even for an untrained AV model, its SR is already quite high (62\%). Therefore, using random sampling of scenarios for training like \textit{SAC w/ rand} and \textit{PER} would inevitably lead to inefficient training, wasting time on scenarios that the AV model does not require training on. 



\begin{figure*}[t]
    \begin{minipage}[t]{0.353\linewidth}
        \begin{subfigure}
            \centering
            \setlength{\abovecaptionskip}{0mm}
            \includegraphics[width=0.98\linewidth]{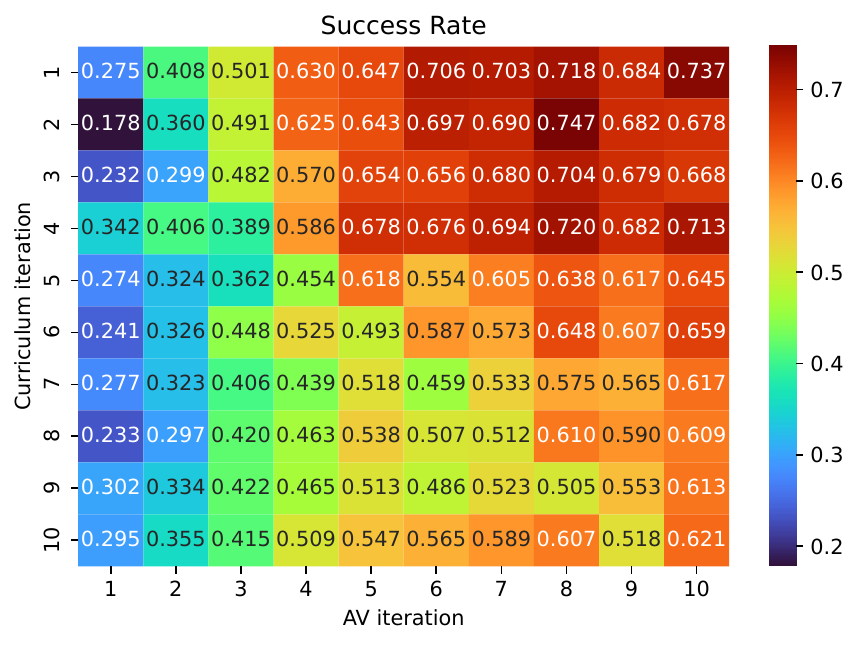}
            \caption{Results of the matrix experiment.}
            \label{fig:heat}
        \end{subfigure}
    \end{minipage}
    \hfill
    \begin{minipage}[t]{0.635\linewidth}
        \begin{subfigure}
            \centering
            \setlength{\abovecaptionskip}{0mm}
            \includegraphics[width=0.98\linewidth]{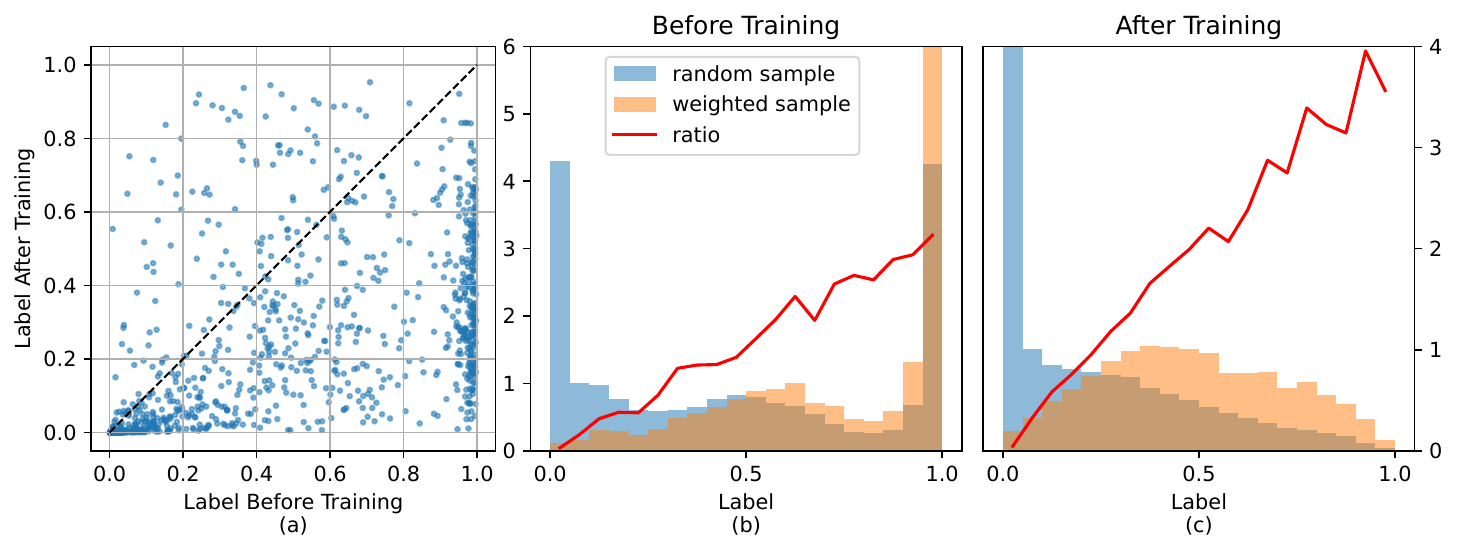}
            \caption{\textbf{(a)} Changes of partial individual scenario labels before and after training.\protect\\ \textbf{(b)(c)} Label distribution before and after training.}
            \label{fig:label}
        \end{subfigure}
    \end{minipage}
\end{figure*}

\subsection{Matrix Experiment}\label{4-2}
In addition to testing the performance change of the AV model before and after training, we also aim to verify the changes in the capabilities of the AV model and the difficulty of scenarios selected by the predictor throughout the training process. Therefore, we conduct the following matrix experiments: we save the AV model and predictor model at each training iteration, and sequentially use each predictor to select scenarios for training curriculum and test them with each AV model. This process yield a $T\times T$ matrix of SR, where $T$ represents the total number of training iterations. For instance, the element ``$AV\ iteration=3,\ Curriculum\ iteration=4$'' represents the testing SR of the AV model trained after 3 iterations when tested on the scenarios selected by the predictor in the 4th iteration. 

As shown in Figure~\ref{fig:heat}, the experimental results demonstrate the following characteristics: 
\begin{itemize}[leftmargin=*,topsep=0pt]
    \item The AV models with a higher number of training iterations achieved higher SR among the scenarios selected by each specific predictor, indicating that the capability of the AV model is indeed continually increasing.
    \item For each specific AV model, there is a decreasing trend in SR among the scenarios selected by subsequent predictors, indicating that the difficulty of the scenarios selected by the predictor is indeed increasing. 
    \item For the AV model in the first training iteration, it exhibits a higher SR on scenarios selected by some of subsequent predictors. This is primarily due to the lower capability of the AV model at this stage, rendering the difficulty level of predictions by the predictors insignificant. This observation further emphasizes the importance of individualized curriculum design in CLIC.
\end{itemize}

\subsection{Analyses on Scenario Reweighting}\label{4-3}
To showcase the changes in the distribution of scenario weights before and after training, we tracked the changes in predicted label values for individual scenarios before and after training. These changes are presented as a scatter plot (Figure~\ref{fig:label}(a)). The scenarios in the figure are concentrated in two regions: First, scenarios with initially small label values tend to be concentrated around 0 after training; Second, scenarios with label values close to 1 before training show varying degrees of decrease in label values after training. Additionally, the majority of scenarios fall below the diagonal line $y=x$, indicating that these scenarios indeed become simpler for the AV model after training. This observation further confirms the conclusion drawn in Section~\ref{4-2}. 

Moreover, to further illustrate the overall scenario reweighting, we also plot histograms of the label distributions obtained through random sampling and weighted sampling before and after training, as shown in Figure~\ref{fig:label}(b)(c). It can be observed that before training, the predicted labels are concentrated at both ends of the interval, while the rest of the distribution appears relatively uniform. However, with weighted sampling, the labels are concentrated near 1, indicating the selection of harder scenarios for the current AV model. After training, as the AV model becomes sufficiently powerful, the predicted labels near 1 significantly decrease, and more labels concentrate near 0. Nonetheless, in order to select relatively hard training scenarios, weighted sampling still acquires a considerable number of scenarios with larger labels. To provide a more intuitive demonstration of the weighted sampling, we also plot the ratio curve of the normalized frequencies corresponding to the two sampling methods. As expected, the ratio curve maintains a proportional relationship for each sampling instance, which is consistent with the setting of Equation~\ref{weighted_sample}. 



\begin{figure}[t]
    \centering
    \setlength{\abovecaptionskip}{0.cm}
    \includegraphics[width=0.75\linewidth]{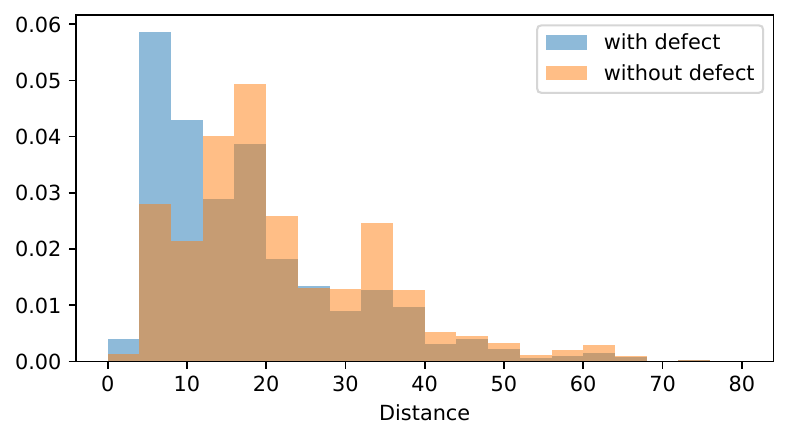}
    \caption{Distribution of distances between the AV and the BVs on its left front side.}
    \vspace{-2mm}
    \label{fig:bv_dis}
\end{figure}

\subsection{Analyses on Individualization}\label{4-4}
To demonstrate that CLIC can indeed generate individualized curricula that identify specific defects in AV model and select more targeted training scenarios accordingly, we intentionally disable a trained AV model to get less aware of BVs on the left front side.
We then let this disabled AV go through CLIC pipeline and compare the selected scenarios with the ones selected for the AV model without defects. 
We analyze the distribution of distances between the AV and the BVs on its left front side in these scenarios illustrated in Figure~\ref{fig:bv_dis}.
As expected, these BVs are positioned closer to the AV, increasing the risk of collisions. 
Furthermore, CLIC selected a significantly higher proportion (\textbf{27.87\%}) of scenarios for AVs with defect where BVs are positioned on the left front side, surpassing the selected proportion (\textbf{17.19\%}) of the AV model with no defect by a substantial margin.
These clearly echo our intuition of providing individualized curricula that target specific defects in training AV.

\section{Conclusion and Future Work}\label{5}
In this paper, we develop a scenario-based continual driving policy optimization framework with Closed-Loop Individualized Curricula (CLIC) technique, composed of three sub-modules for flexible implementation choices: AV Evaluation, Scenario Selection and AV Training.
CLIC approaches AV Evaluation as a difficulty prediction task by training a discriminator on AV collision labels to estimate the potential failure probability of AV within corresponding scenarios.
With the prediction results from the discriminator, CLIC reweights scenarios to select individualized curricula for AV training that incorporate less easy cases with more challenging ones according to current AV capability.
To summarize, CLIC not only fully exploits the vast historical scenario library for closed-loop AV training rather than just for AV testing, but also facilitates AV improvement by individualizing its training on more helpful scenarios that match its current capability out of the diverse yet poorly organized library.
Through extensive experimental analyses, CLIC outperforms other competing curriculum-based baselines, especially in handling safety-critical scenarios, while minimizing the degradation of performance in regular scenarios. 
For future work, we are interested in exploring scenario libraries with more complex road topologies and other types of traffic participants, as well as investigating theoretical guarantees for AV improvement within these individualized curricula.
We also believe that CLIC is not only applicable to the field of autonomous driving policy optimization, but also a general method for complex tasks and continual learning in robotics.





\section*{Acknowledgement}
This work is supported by National Natural Science Foundation of China under Grant No. 62333015 and Beijing Natural Science Foundation L231014. 

\bibliographystyle{IEEEtran}
\bibliography{mylib}


\clearpage
\appendix
\subsection{Dataset Analysis and Visualization}\label{appendix_dataset}
\begin{figure}[H]
    \centering
    \includegraphics[width=0.98\linewidth]{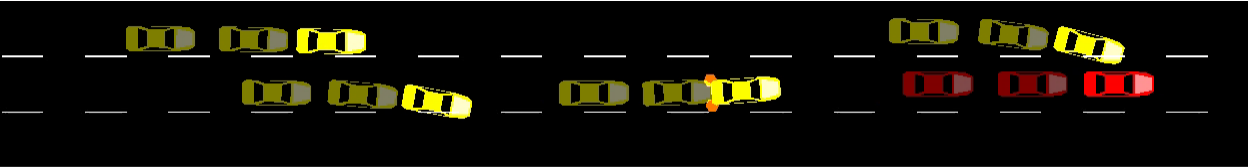}
    \caption{An illustrated example of the scenarios.}
    \label{fig:scenario_example}
\end{figure}

All of the scenario data we use is generated by (Re)$^2$H2O~\cite{niu2023re} and saved in the form of static vectors after preprocessing according to the definition in Section~\ref{3-1-1}, making it convenient to flexibly load into the data stream for reproduction when needed. 
This scenario library records $M=65494$ extreme scenarios (excluding scenarios where BVs collided with each other already) formed when 1$\sim$4 BVs attempted to attack AV on a 3-lane straight highway with a length of about 200m, with a sampling interval of $\Delta t=0.04{\rm s}$. Among them, there are 53040 scenarios where AV had accidents, with an accident rate of 81.0\%. Figure~\ref{fig:scenario_example} shows an illustrated example of the scenarios where AV has an accident with one BV. 

During scenario generation, certain limitations were imposed on the actions of BVs to meet common vehicle dynamics constraints. The acceleration of BVs is limited to $[-7.84,5.88]({\rm m/s^2})$ (i.e. $[-0.8g,0.6g]$, $g$ is the gravitational acceleration, taken as $9.8{\rm m/s^2}$), while the angular velocity is limited to $[-\pi/3,\pi/3]({\rm rad/s})$. In addition, the maximum velocity limit on the road is $40{\rm m/s}$ and negative velocity (i.e. reversing) is not allowed. 

\begin{figure}[H]
    \centering
    \begin{minipage}[H]{0.32\linewidth}
        \begin{subfigure}
            \centering
            \setlength{\abovecaptionskip}{0mm}
            \includegraphics[width=0.97\linewidth]{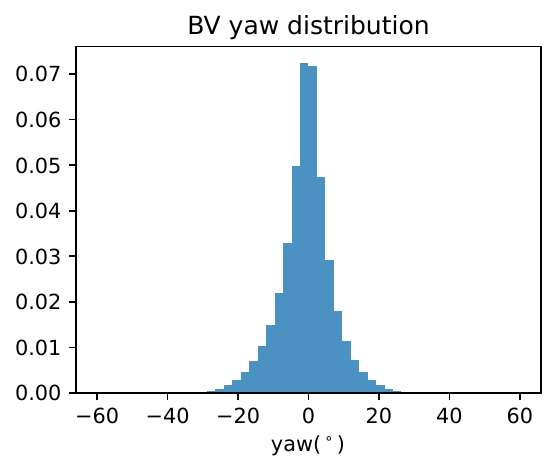}
            \label{fig:bv_yaw}
        \end{subfigure}
    \end{minipage}
    \hfill
    \begin{minipage}[H]{0.32\linewidth}
        \begin{subfigure}
            \centering
            \setlength{\abovecaptionskip}{0mm}
            \includegraphics[width=0.97\linewidth]{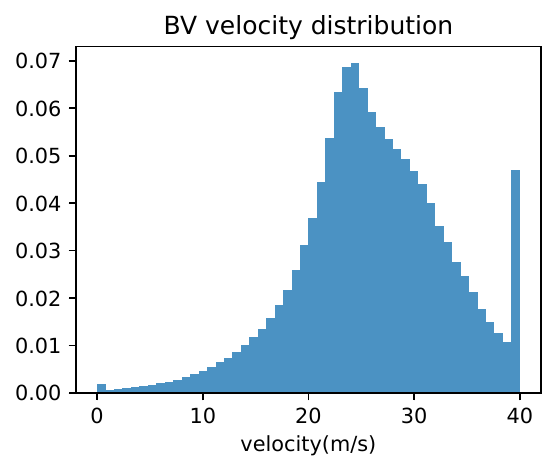}
            \label{fig:bv_vel}
        \end{subfigure}
    \end{minipage}
    \hfill
    \begin{minipage}[H]{0.32\linewidth}
        \begin{subfigure}
            \centering
            \setlength{\abovecaptionskip}{0mm}
            \includegraphics[width=0.97\linewidth]{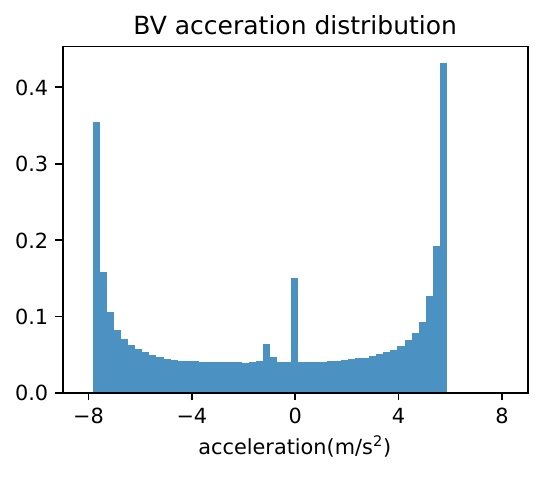}
            \label{fig:bv_acc}
        \end{subfigure}
    \end{minipage}
    \vspace{-3mm}
    \caption{The \textbf{yaw}, \textbf{velocity} and \textbf{acceleration} distribution of all BVs.}
    \label{fig:bv_distribution}
    \vspace{-2mm}
\end{figure}

\begin{figure}[H]
    \centering
    \includegraphics[width=0.99\linewidth]{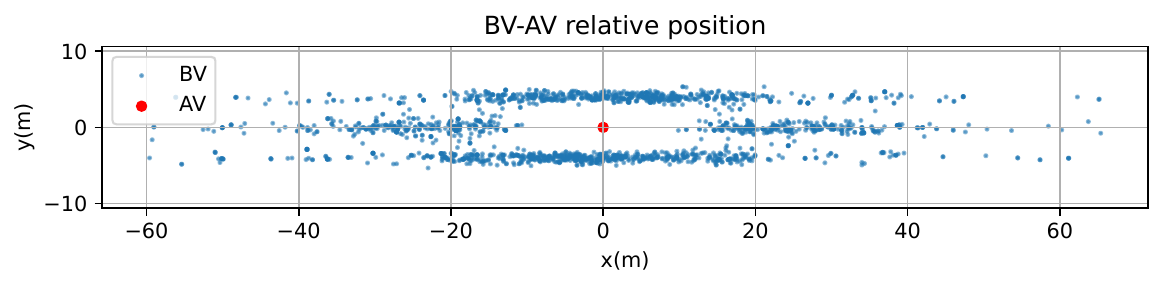}
    \caption{The initial position of partial BV relative to AV.}
    \label{fig:bv_av_pos}
\end{figure}

In order to better illustrate some of the features of the scenario library, we have calculated the distribution of the following features and visualized them: the yaw, velocity and acceleration of BVs, the distance between BVs, the initial distance between BVs and AV, and the initial position of BV relative to AV. Figure~\ref{fig:bv_distribution}~$\sim$~\ref{fig:bv_av_pos} show these features respectively. 

\begin{figure}[H]
    \begin{minipage}[H]{0.48\linewidth}
        \begin{subfigure}
            \centering
            \setlength{\abovecaptionskip}{0mm}
            \includegraphics[width=0.98\linewidth]{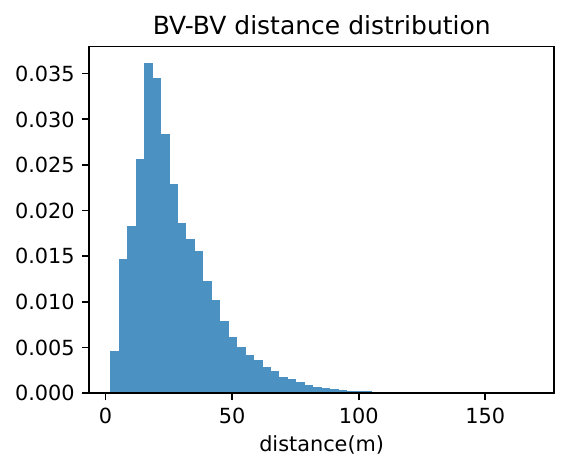}
            \label{fig:bv_bv_dis}
        \end{subfigure}
    \end{minipage}
    \hfill
    \begin{minipage}[H]{0.48\linewidth}
        \begin{subfigure}
            \centering
            \setlength{\abovecaptionskip}{0mm}
            \includegraphics[width=0.98\linewidth]{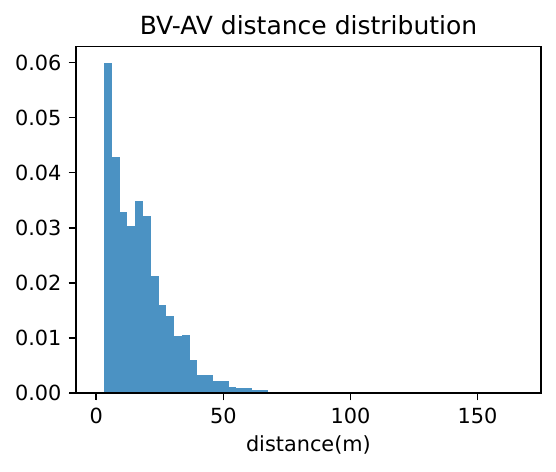}
            \label{fig:bv_av_dis}
        \end{subfigure}
    \end{minipage}
    \vspace{-3mm}
    \caption{\textbf{Left}: The distance distribution between BVs. \\ \textbf{Right}: The initial distance distribution between BVs and AV.}
    \label{fig:distance_distribution}
\end{figure}

From the above feature distribution statistics, it can be seen that although the scenario library contains a large number of scenarios where AV accidents occur during generation, there are also many scenarios where BVs' driving behavior patterns are quite conventional, and the BVs are relatively sparse. Therefore, it is highly likely that these scenarios overfitted the AV model $\mu(\mathbf{a}|\mathbf{s})$ at the time of generation and are not applicable to another AV model $\pi_{\phi}(\mathbf{a}|\mathbf{s})$, resulting in high SR in our testing experiment. 
In addition, these feature distributions also reflect the diversity of the scenario library data, making them more suitable for our algorithmic framework that utilizes closed-loop feedback for individualized training scenario selection. 

\subsection{Implementation Details}\label{appendix_implementation}
Due to the fact that conventional neural networks must have fixed dimension of inputs, but the different number of BVs and timesteps in the scenarios can result in inconsistent input dimensions, we have to initialize the neural network by obtaining the maximum possible input dimension in advance, and fill in the remaining dimensions with zeros when the input dimension is insufficient. 
In this way, the actual input dimension of the predictor is $6\times(1+4\times100)=2406$ (Each vehicle's data at each moment contains 6 dimensions, including timestep $t$, vehicle ID, $x_t^{(i)}$, $y_t^{(i)}$, $v_t^{(i)}$, $\theta_t^{(i)}$; all scenario data contain up to 4 BVs, with a maximum timestep of 100; 1 represents the AV information at the initial moment), while the actual input state space dimension of the AV model is $4\times(1+4)=20$ (The state of each vehicle includes 4 dimensions: $x_t^{(i)}$, $y_t^{(i)}$, $v_t^{(i)}$, $\theta_t^{(i)}$, with a total of 1 AV and no more than 4 BVs). On the other hand, the object of action space is only the AV itself, so its dimension is determined to be 2 ($\Delta v_t^{(i)}$, $\Delta\theta_t^{(i)}$) according to Section~\ref{3-1-1}. 

Table~\ref{table_reward_coefficient} lists the values of the coefficients corresponding to each reward function item, which have been adjusted through certain experiments, as well as the AV speed range in the reward function based on road speed limit. 

\begin{table}[htbp]
  \centering
  \caption{Coefficients and AV speed range of the reward function}
  \small
    \begin{tabular}{clc}
    \toprule
    Coefficient & \multicolumn{1}{>{\centering}c}{Description} & Value \\
    \midrule
    $\rho_{acc}$ & Coefficient of the accident reward item & 40 \\
    $\rho_{vel}$ & Coefficient of the velocity reward item & 0.8 \\
    $\rho_{yaw}$ & Coefficient of the yaw reward item & $6/\pi$ \\
    $\rho_{lane}$ & Coefficient of the lane reward item & 2.0 \\
    $v_{max}$ & The maximum speed of AV & 40${\rm m/s}$ \\
    $v_{min}$ & The minimum speed of AV & 0${\rm m/s}$ \\
    \bottomrule
    \end{tabular}
  \label{table_reward_coefficient}
\end{table}

The network structure and hyperparameters of our algorithm are listed in Table~\ref{table_hyperparameters}. Except for the temperature parameter $\alpha=0.1$ and predictor training epochs 20 which are adjusted through experiments, most of these hyperparameters are based on empirical values. 

\begin{table*}[htbp]
  \centering
  \caption{Hyperparameters}
  \normalsize
    \begin{tabular}{m{50mm}m{100mm}c}
    \toprule
    \multicolumn{1}{>{\centering}m{50mm}}{Hyperparameter} & \multicolumn{1}{>{\centering}m{100mm}}{Description} & Value \\
    \midrule
    Evaluate size $m$ & Number of scenarios tested in AV Evaluation stage & 4096 \\
    Train size $n$ & Number of scenarios trained in AV Training stage & 128 \\
    Layer number & Number of hidden layers in each network (all) & 3 \\
    Hidden size & Number of hidden units in each hidden layer & 256 \\
    Batch size & Number of examples in each batch when training the predictor & 128 \\
    Learning rate $\lambda$ & The step size in network gradient descent (all) & 10$^{-4}$ \\
    Optimizer & The optimizer used in gradient descent (all) & Adam \\
    Nonlinearity & Nonlinear activation functions in networks & ReLU \\
    Nonlinearity (predictor output) & Nonlinear activation function in the predictor output layer & Softmax \\
    Nonlinearity (policy output) & Nonlinear activation function in the output layer of the policy net& Tanh \\
    Iterations $T$ & The number of iterations of the entire closed-loop algorithm & 10 \\
    Epochs & The number of training epochs when training the predictor & 20 \\
    Episodes & The number of training episodes on each scenario & 10 \\
    Temperature parameter $\alpha$ & Entropy regularization coefficient in SAC objective function & 0.1 \\
    Memory capacity & The maximum capacity of the replay buffer in SAC & 10$^6$ \\
    Discount factor $\gamma$ & The future reward discount factor in RL & 0.99 \\
    Target smoothing coefficient $\tau$ & Update coefficient for target network parameter soft update & 0.01 \\
    \bottomrule
    \end{tabular}
  \label{table_hyperparameters}
\end{table*}

All of our experiments are conducted on a Linux server equipped with an 8-core Intel(R) Xeon(R) CPU E5-2620 v4 and 3 NVIDIA Geforce GTX 1080Ti chips. Under the above experimental setup, the entire closed-loop training process takes about 6 hours, with the \textit{AV training} stage in Section~\ref{3-3-3} accounting for an average of over \textbf{83.0\%} of the time. This indicates that our individualized curriculum generation process does not bring significant time costs and computational burden. 

\subsection{Additional Experiment Results}\label{appendix_additional}
\subsubsection{Additional Ablations, Baselines and Metrics}
In addition to the baselines mentioned in Section~\ref{4-1-1-2}, our experiment also includes the following experimental settings for comparison: 
\begin{itemize}[leftmargin=*,topsep=0pt]
    \item \textbf{\textit{alpha\_0.15}} and \textbf{\textit{alpha\_0.2}}: In Table~\ref{table_hyperparameters}, we set the temperature parameter $\alpha$ in SAC as 0.1, and in fact, we also try $\alpha=0.15$ and $\alpha=0.2$ for the experiment and compare the results. The larger the value of $\alpha$, the more inclined the model is to explore policy diversity. 
    \item \textbf{\textit{auto\_alpha}}: As shown in~\cite{haarnoja2018softarxiv}, the temperature parameter $\alpha$ can also be designed with a loss function and automatically adjusted by gradient decent: 
    \begin{equation}
            J(\alpha)=\mathbb{E}_{\mathbf{a}_t\sim\pi_t}[-\alpha\log\pi_t(\mathbf{a}_t|\mathbf{s}_t)-\alpha\Bar{\mathcal{H}}]
    \end{equation}
    \begin{equation}
        \alpha\leftarrow\alpha-\lambda\hat{\nabla}_{\alpha}J(\alpha)
    \end{equation}
    where $\Bar{\mathcal{H}}$ is the target entropy, and $\lambda$ is the learning rate. Here $\Bar{\mathcal{H}}$ is set as 0, and $\lambda$ remains consistent with others, i.e. 10$^{-4}$. 
    \item \textbf{\textit{dropout}}: Add dropout~\cite{srivastava2014dropout} during predictor training to prevent overfitting, and set the dropout probability as 0.2. 
    \item \textbf{\textit{batch\_1:1}}: In order to prevent the serious imbalance of positive and negative examples during the predictor training process from leading to poor predictor performance, the number of positive and negative examples in each batch is ensured to be 1:1 by copying examples with the fewer labels during the predictor training process. 
    \item \textbf{\textit{order}}: The algorithm settings are basically the same as CLIC, but when selecting training scenarios, they are not weighted sampled based on the predicted difficulty labels. Instead, the scenarios are sorted in order according to their labels, and then all scenarios are divided into a specific number of intervals according to the order. From each interval we randomly sample one scenario to form the training scenarios. 
    \item \textbf{\textit{PCL\_label}}: In addition to using the number of BV as a difficulty classification basis for PCL, the initial predicted difficulty labels can also be divided into different intervals for PCL, called \textit{PCL\_label}, and correspondingly, the baseline \textit{PCL} in Section~\ref{4-1-1-2} is called \textit{PCL\_bv\_num} in Table~\ref{table_complete_results_safety} and Table~\ref{table_complete_results_efficiency_comfort}. Here, all scenarios are divided into 5 curriculum stages at intervals of 0.2 (The range of label is $[0,1]$), with 2 iterations of training for each stage. 
\end{itemize}
In addition to the safety related metrics mentioned in Section~\ref{4-1-1-3}, we also consider more metrics related to efficiency and comfort: 
\begin{itemize}[leftmargin=*,topsep=0pt]
    \item \textbf{vel} (${\rm m/s}$): The average velocity of AV in all scenarios during testeing. 
    \item \textbf{succ\_vel} (${\rm m/s}$): The average velocity of AV only in successful scenarios during testing, reflecting the efficiency of AV while ensuring safety. 
    \item \textbf{acc} (${\rm m/s^2}$): The average acceleration of AV in all scenarios during testing. 
    \item \textbf{jerk} (${\rm m/s^3}$): The average derivative of acceleration over time of AV in all scenarios during testeing. 
    \item \textbf{ang\_vel} (${\rm rad/s}$): The average angular velocity of AV in all scenarios during testing. 
    \item \textbf{lat\_acc} (${\rm m/s^2}$): The average lateral acceration of AV in all scenarios during testing. 
\end{itemize}
Among them, the first two metrics reflect the efficiency of AV and are expected to be larger, while the last four metrics reflect the comfort of AV and are expected to be smaller. 

\subsubsection{Complete Results of Comparative and Ablation Experiments}
Adding the above experimental conditions or baselines and metrics, the complete results of the comparative experiments are shown in Table~\ref{table_complete_results_safety} and Table~\ref{table_complete_results_efficiency_comfort}, which are safety related metrics and efficiency \& comfort related metrics respectively. 

\begin{table*}[htbp]
  \centering
  \caption{Complete results of comparetive and ablation experiments \\(Safety related metrics)}
  \small
  \renewcommand{\arraystretch}{1.0}
    \begin{tabular}{c|ccccc}
    \toprule
     & SR $\uparrow$ & FNR $\uparrow$ & TNR $\uparrow$ & CPS $\downarrow$ & CPM $\downarrow$ \\
     \midrule
    untrained & {61.99$\pm$4.03} & {---} & {---} & {0.301$\pm$0.044} & {0.0112$\pm$0.0017} \\
    \midrule
    \textbf{CLIC} & {\textbf{79.76$\pm$1.67}} & {\textbf{62.3$\pm$3.1}} & {\textbf{90.3$\pm$1.6}} & {\textbf{0.147$\pm$0.015}} & {0.0050$\pm$0.0005} \\
    (alpha\_0.15) & {76.45$\pm$1.68} & {60.8$\pm$2.8} & {85.8$\pm$3.7} & {0.177$\pm$0.015} & {0.0055$\pm$0.0004} \\
    (alpha\_0.2) & {78.84$\pm$0.97} & {61.4$\pm$2.9} & {89.2$\pm$1.6} & {0.148$\pm$0.024} & {\textbf{0.0049$\pm$0.0009}} \\
    (auto\_alpha) & {77.66$\pm$2.58} & {59.9$\pm$4.9} & {88.1$\pm$3.1} & {0.165$\pm$0.024} & {0.0053$\pm$0.0007} \\
    (dropout) & {76.52$\pm$3.05} & {58.7$\pm$5.6} & {87.5$\pm$1.7} & {0.179$\pm$0.027} & {0.0060$\pm$0.0014} \\
    (batch\_1:1) & {78.34$\pm$3.22} & {63.0$\pm$4.9} & {87.6$\pm$1.6} & {0.160$\pm$0.027} & {0.0055$\pm$0.0010} \\
    (order) & {76.07$\pm$1.63} & {57.0$\pm$2.9} & {87.6$\pm$2.8} & {0.175$\pm$0.022} & {0.0058$\pm$0.0006} \\
    \midrule
    SAC w/ rand & {75.35$\pm$2.73} & {55.5$\pm$4.6} & {87.4$\pm$2.9} & {0.187$\pm$0.025} & {0.0060$\pm$0.0008} \\
    SAC w/ rand+fail & {76.40$\pm$1.09} & {58.4$\pm$2.4} & {87.2$\pm$1.5} & {0.177$\pm$0.009} & {0.0058$\pm$0.0004} \\
    SAC w/ fail & {77.53$\pm$2.47} & {60.3$\pm$3.8} & {87.9$\pm$2.3} & {0.165$\pm$0.023} & {0.0059$\pm$0.0014} \\
    PCL\_bv\_num & {76.24$\pm$1.88} & {58.0$\pm$3.7} & {87.4$\pm$2.6} & {0.175$\pm$0.016} & {0.0058$\pm$0.0005} \\
    PCL\_label & {72.99$\pm$2.33} & {53.2$\pm$4.3} & {85.0$\pm$3.2} & {0.207$\pm$0.021} & {0.0072$\pm$0.0017} \\
    PER & {74.18$\pm$4.05} & {55.1$\pm$2.8} & {86.0$\pm$4.7} & {0.194$\pm$0.036} & {0.0075$\pm$0.0014} \\
    \bottomrule
    \end{tabular}
  \label{table_complete_results_safety}
\end{table*}

\begin{table*}[htbp]
  \centering
  \caption{Complete results of comparetive and ablation experiments \\(Efficiency \& Comfort related metrics)}
  \small
  \renewcommand{\arraystretch}{1.0}
    \begin{tabular}{c|cc|cccc}
    \toprule
     & vel $\uparrow$ & succ\_vel $\uparrow$ & acc $\downarrow$ & jerk $\downarrow$ & ang\_vel $\downarrow$ & lat\_acc $\downarrow$ \\
     \midrule
    untrained & 27.01$\pm$0.31 & 26.88$\pm$0.72 & 0.61$\pm$0.19 & 0.18$\pm$0.05 & 0.028$\pm$0.013 & 0.76$\pm$0.36 \\
    \midrule
    \textbf{CLIC} & 28.92$\pm$0.92 & 29.32$\pm$1.05 & 4.06$\pm$0.24 & 5.70$\pm$0.51 & 0.229$\pm$0.015 & 6.41$\pm$0.47 \\
    (alpha\_0.15) & \textbf{31.96$\pm$0.98} & \textbf{32.67$\pm$1.02} & 4.15$\pm$0.43 & 3.87$\pm$1.17 & 0.206$\pm$0.009 & 6.16$\pm$0.33 \\
    (alpha\_0.2) & 30.18$\pm$0.86 & 30.78$\pm$1.09 & \textbf{3.55$\pm$0.14} & 3.68$\pm$0.40 & 0.188$\pm$0.013 & 5.37$\pm$0.35 \\
    (auto\_alpha) & 31.13$\pm$0.53 & 31.57$\pm$0.76 & 3.92$\pm$0.08 & 4.61$\pm$0.38 & 0.201$\pm$0.008 & 5.88$\pm$0.27 \\
    (dropout) & 30.50$\pm$2.11 & 30.82$\pm$2.40 & 4.38$\pm$0.30 & 4.77$\pm$1.33 & 0.222$\pm$0.014 & 6.32$\pm$0.34 \\
    (batch\_1:1) & 29.33$\pm$1.84 & 30.04$\pm$2.08 & 4.28$\pm$0.32 & 5.09$\pm$0.67 & 0.225$\pm$0.008 & 6.32$\pm$0.45 \\
    (order) & 30.48$\pm$1.74 & 31.37$\pm$1.93 & 4.28$\pm$0.25 & 4.32$\pm$0.35 & 0.219$\pm$0.017 & 6.21$\pm$0.67 \\
    \midrule
    SAC w/ rand & 31.42$\pm$1.11 & 31.96$\pm$1.30 & 4.12$\pm$0.29 & 3.74$\pm$0.56 & 0.202$\pm$0.011 & 5.91$\pm$0.43 \\
    SAC w/ rand+fail & 30.78$\pm$1.77 & 31.33$\pm$2.06 & 3.97$\pm$0.17 & 4.37$\pm$0.61 & 0.203$\pm$0.016 & 5.94$\pm$0.50 \\
    SAC w/ fail & 28.60$\pm$2.61 & 29.07$\pm$3.01 & 3.74$\pm$0.14 & 4.87$\pm$0.73 & 0.201$\pm$0.011 & 5.60$\pm$0.53 \\
    PCL\_bv\_num & 30.21$\pm$1.77 & 30.76$\pm$1.61 & 4.02$\pm$0.31 & \textbf{3.32$\pm$0.24} & 0.185$\pm$0.011 & 5.26$\pm$0.31 \\
    PCL\_label & 29.50$\pm$3.63 & 30.36$\pm$4.08 & 4.05$\pm$0.65 & 4.38$\pm$0.67 & 0.200$\pm$0.010 & 5.65$\pm$0.63 \\
    PER & 26.20$\pm$3.01 & 26.47$\pm$3.34 & 3.92$\pm$0.42 & 4.38$\pm$0.58 & \textbf{0.174$\pm$0.012} & \textbf{4.60$\pm$0.77} \\
    \bottomrule
    \end{tabular}
  \label{table_complete_results_efficiency_comfort}
  \vspace{-1mm}
\end{table*}

The experimental results indicate that the CLIC method (including fine-tuning $\alpha$) has achieved the best performance in various safety and efficiency related metrics. When $\alpha=0.15$, the efficiency of AV is the highest, and when $\alpha=0.2$, AV performs quite well in terms of comfort. 
In addition, thanks to the velocity reward item in our reward function (Equation~\ref{reward_vel}), the AV models trained by various methods have made gratifying progress in efficiency compared to untrained AV model. 
However, improvements in the training process of the predictor are not reflected in the final results, and the loss function of the predictor itself also dose not significantly decrease further after these improvements. 
The method of selecting training scenarios in order of difficulty labels is indeed not as good as weighted sampling. 
Furthermore, as mentioned in Section~\ref{4-1-2}, PCL has not fully utilized the powerful advantages of curriculum learning, as its predefined curriculum settings completely ignore the ability feedback of the current AV model, thus lacking adaptability. 

It is also worth mentioning that although the untrained AV model has shown seemingly good SR, they have exposed their practical shortcomings in CPS and CPM, more than twice as high as CLIC. Through comfort-related metrics, it can also be seen that the untrained AV model is difficult to take action to avoid accidents and only engages in trivial driving in a na\"ive manner. This also reflects the importance of individualized curricula, otherwise these once challenging scenarios would lose their transferability and the value of reuse even in such a na\"ive model. 

\subsection{Experiments for Predictor}\label{appendix_predictor}
In order to verify the performance of the predictor in CLIC, as well as the reasons for selecting the model structure and hyperparameters of the predictor, we conduct independent experiments on the predictor. We rollout all scenarios with an untrained AV model and get a set of fixed accident labels for the training and testing of predictor using supervised learning. This set of labels contains 40347 accident scenarios (label = 1) and 25147 accident free scenarios (label = 0), with the positive and negative sample ratio of 1.60. In order to be consistent with the actual use in CLIC, only 4096 (i.e. \textit{Evaluate size} in Table~\ref{table_hyperparameters}) scenarios and their labels are divided into training set in all 65494 scenarios, and all the remaining scenarios are regarded as validation set (test set). Such a large proportion of test samples and training samples will inevitably lead to imperfect performance of the predictor on the test set. 

In our experiment, we choose Multi-Layer Perceptron (MLP)~\cite{rumelhart1986learning}, Recurrent Neural Network (RNN) and Long Short Term Memory (LSTM)~\cite{hochreiter1997long} as the alternative models of the predictor. The reason for choosing the latter two is that the scenario itself has a strong time sequence properties. Therefore, when propagating forward, we first input the initial AV state into an embedding layer to obtain the initial hidden state, and then input the BV state of each timestep as a timing input. Finally, the output passes through a fully connected layer and \texttt{Softmax} activation function to achieve classification. In terms of evaluation metrics, we use BCE Loss (Equation~\ref{bceloss}) and classification accuracy on the test set to measure the performance of the predictor. For the sake of uniformity, the number of hidden layer units of all models is consistent, i.e. \textit{hidden\_size} in Table~\ref{table_hyperparameters}. All the experiments are conducted on 5 random seeds. 

\begin{figure}[H]
    \centering
    \includegraphics[width=0.99\linewidth]{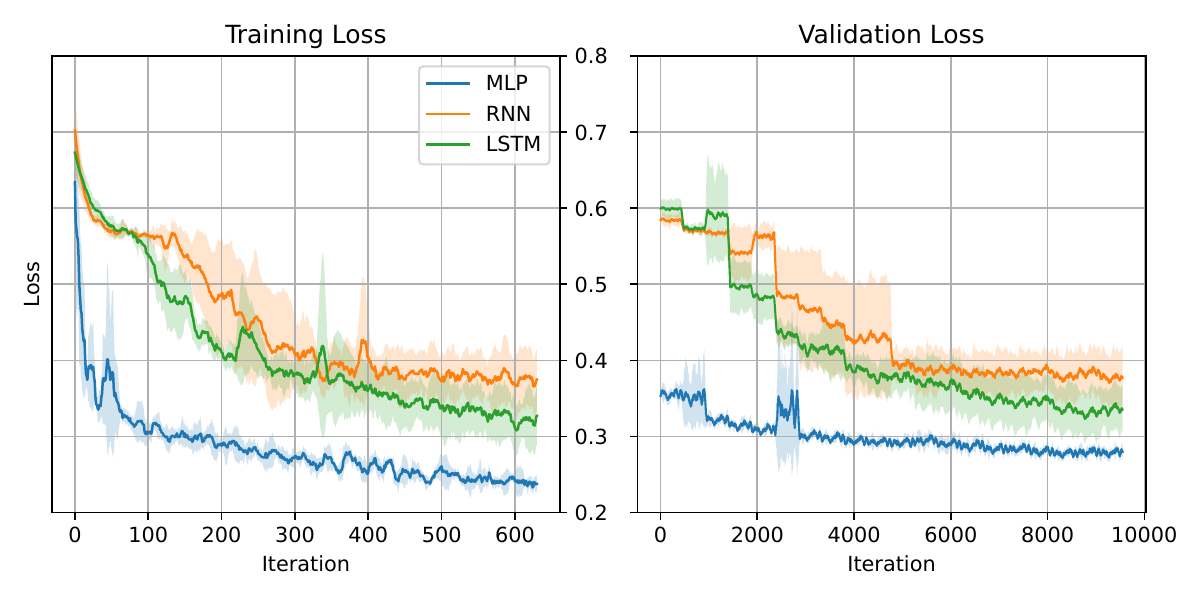}
    \caption{The training and validation loss curve of different predictor models.}
    \label{fig:predictor_loss}
\end{figure}

In Figure~\ref{fig:predictor_loss}, we draw the loss curve of the three predictor models on the training set and the validation set, and process them by Exponential Moving Average (EMA) to obtain better visual effect. The abscissa is the number of iterations, which are $\lceil4096\div128\rceil\times20=640$ and $\lceil(65494-4096)\div128\rceil\times20=9600$ respectively (i.e. $\lceil$\textit{Evaluation size} $\div$ \textit{Batch size}$\rceil\times$ \textit{Epochs} and $\lceil (M-$\textit{Evaluation size}$)\ \div$ \textit{Batch size}$\rceil\times$ \textit{Epochs}). 
It can be found from the figure that the performance of MLP is always better than RNN and LSTM in both the training set and the validation set, with faster convergence speed and less loss. In addition, the variance of MLP is also smaller than the other two, indicating that it has stronger stability and robustness. The structure of RNN and LSTM are very similar, so the change trend of training curve is roughly the same. Due to the gating mechanism added to RNN to prevent forgetting, LSTM performs slightly better than RNN on both dataset. 

\begin{table}[H]
  \centering
  \caption{Performance on the test set after training and the complexity of different predictor models}
  \small
    \begin{tabular}{c|ccc}
    \toprule
    Model & \textbf{MLP} & RNN & LSTM \\
    \midrule
    Loss $\downarrow$ & \textbf{0.279$\pm$0.004} & 0.377$\pm$0.034 & 0.336$\pm$0.029 \\
    Accuracy $\uparrow$ & \textbf{88.10$\pm$0.28} & 84.50$\pm$1.64 & 85.55$\pm$1.62 \\
    Params $\downarrow$ & 682.5k & \textbf{74.5k} & 291.07k \\
    FLOPs $\downarrow$ & \textbf{682.5k} & 7.25M & 29.14M \\
    \bottomrule
    \end{tabular}
  \label{table_predictor}
\end{table}

In Table~\ref{table_predictor}, we list the loss and classification accuracy of the three predictor models on the test set after training the same number of epochs (i.e. \textit{Epochs} in Table~\ref{table_hyperparameters}), as well as parameter quantities and floating-point operations (FLOPs) related to model complexity. It can be seen that MLP has the least loss and highest classification accuracy on the test set, and has the minimum amount of calculation, that is, the fastest training speed. Although the parameter quantity of MLP is significantly larger than those of the other two because of its fully connected structure, it does not need to save a large number of intermediate variables for recurrent calculation, and the overall memory occupied by MLP is also the smallest. On the other hand, the performance of LSTM is slightly improved compared with RNN, but the corresponding cost is that its complexity is increased by about 4 times, and its computational complexity is more than 40 times that of MLP. 

To sum up, MLP is the best performance, fastest convergence, most robust and lightest weight of the above three models, so it is chosen as the predictor in CLIC. 

\subsection{Qualitative Results}\label{appendix_qualitative}
Through SUMO GUI, we visualize some driving scenarios in the experiments and obtain these qualitative results to support our conclusion. 

\begin{figure*}[h]
    \centering
    \begin{minipage}[H]{0.49\linewidth}
        \begin{subfigure}
            \centering
            \setlength{\abovecaptionskip}{0mm}
            \includegraphics[width=0.95\linewidth]{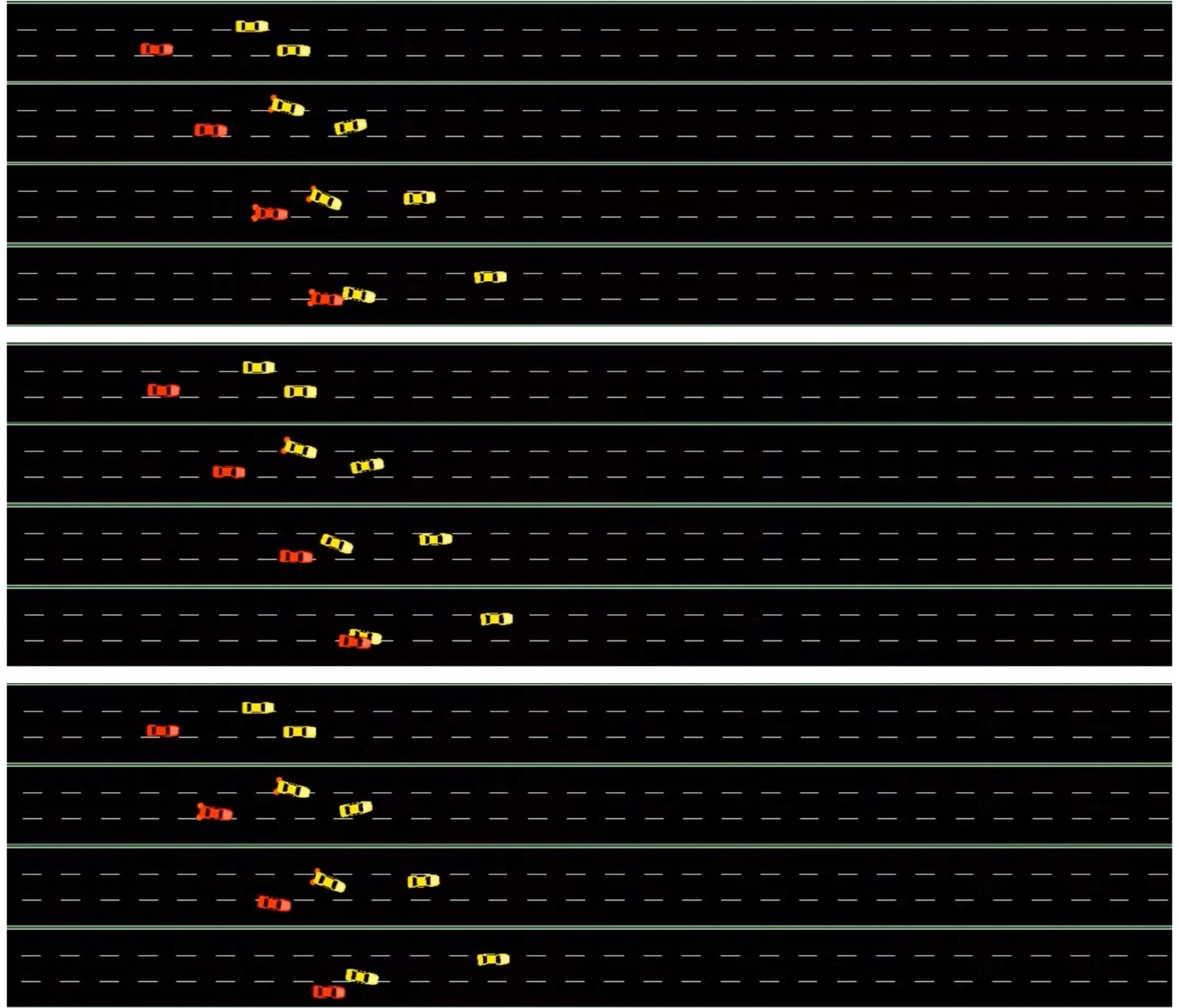}
        \end{subfigure}
    \end{minipage}
    \hfill
    \begin{minipage}[H]{0.49\linewidth}
        \begin{subfigure}
            \centering
            \setlength{\abovecaptionskip}{0mm}
            \includegraphics[width=0.95\linewidth]{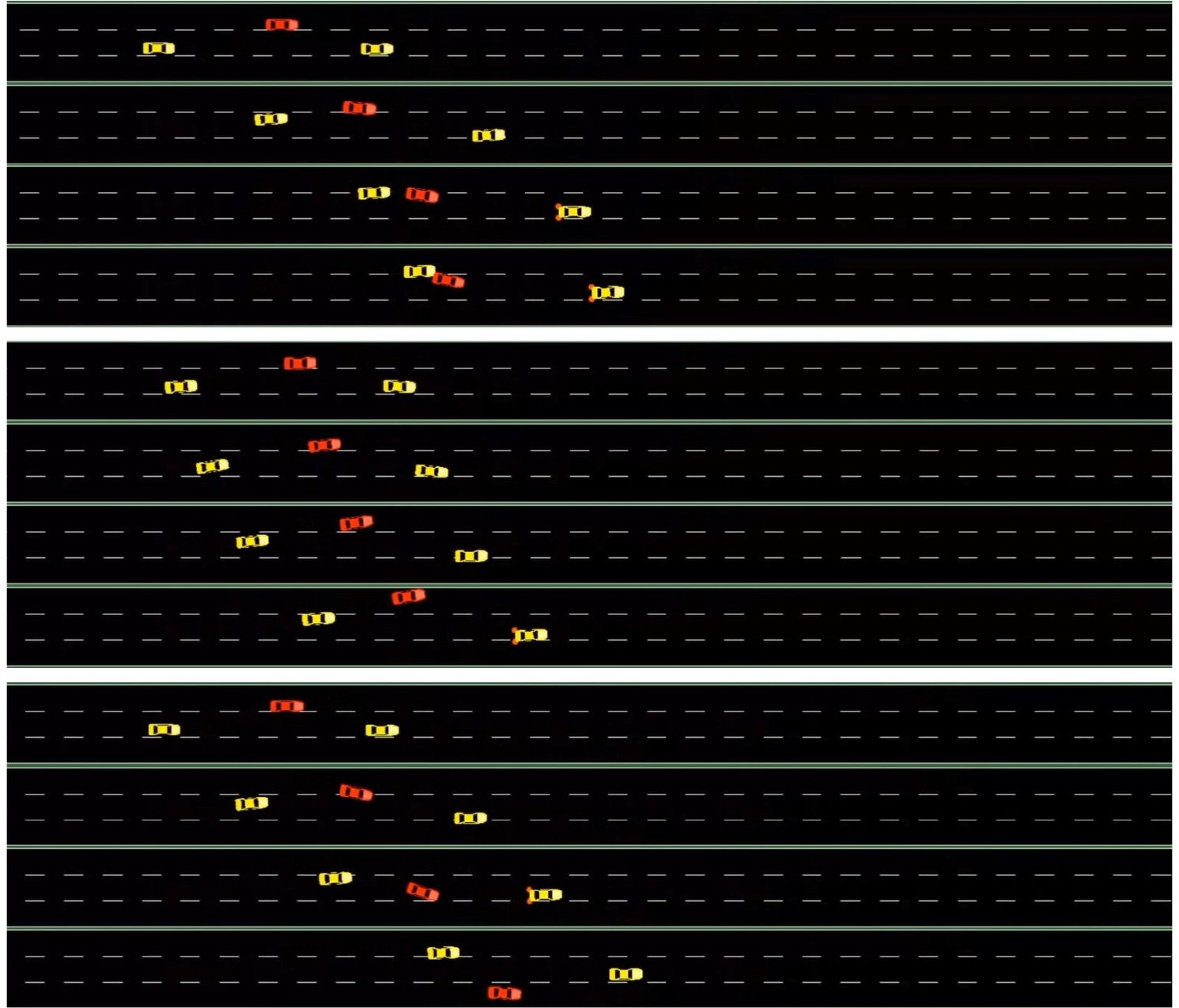}
        \end{subfigure}
    \end{minipage}
    \caption{\textit{Comparative Experiment}: The performance of different AV models in two same scenarios. From top to bottom, they are: \textbf{untrained}, \textbf{trained by \textit{SAC w/ rand}} (no curriculum), and \textbf{trained by CLIC}. Only the AV model trained by CLIC can successfully handle these two scenarios and avoid accidents.}
    \label{fig:comparative_vis}
\end{figure*}

\begin{figure*}[h]
    \centering
    \begin{minipage}[H]{0.5\linewidth}
        \begin{subfigure}
            \centering
            \setlength{\abovecaptionskip}{0mm}
            \includegraphics[width=0.98\linewidth]{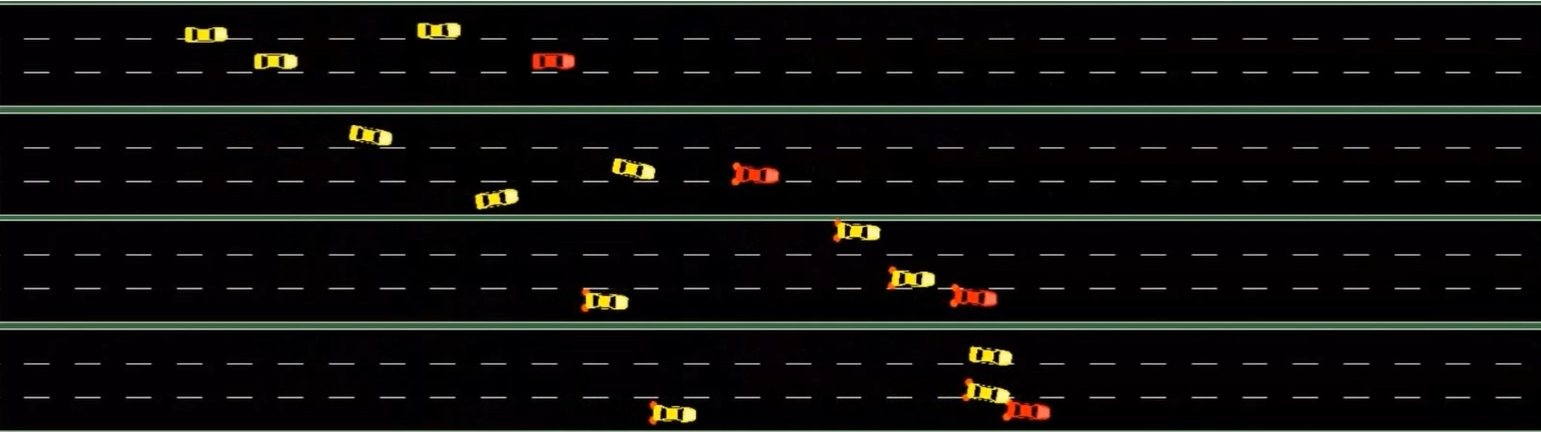}
        \end{subfigure}
    \end{minipage}
    \hfill
    \begin{minipage}[H]{0.5\linewidth}
        \begin{subfigure}
            \centering
            \setlength{\abovecaptionskip}{0mm}
            \includegraphics[width=0.98\linewidth]{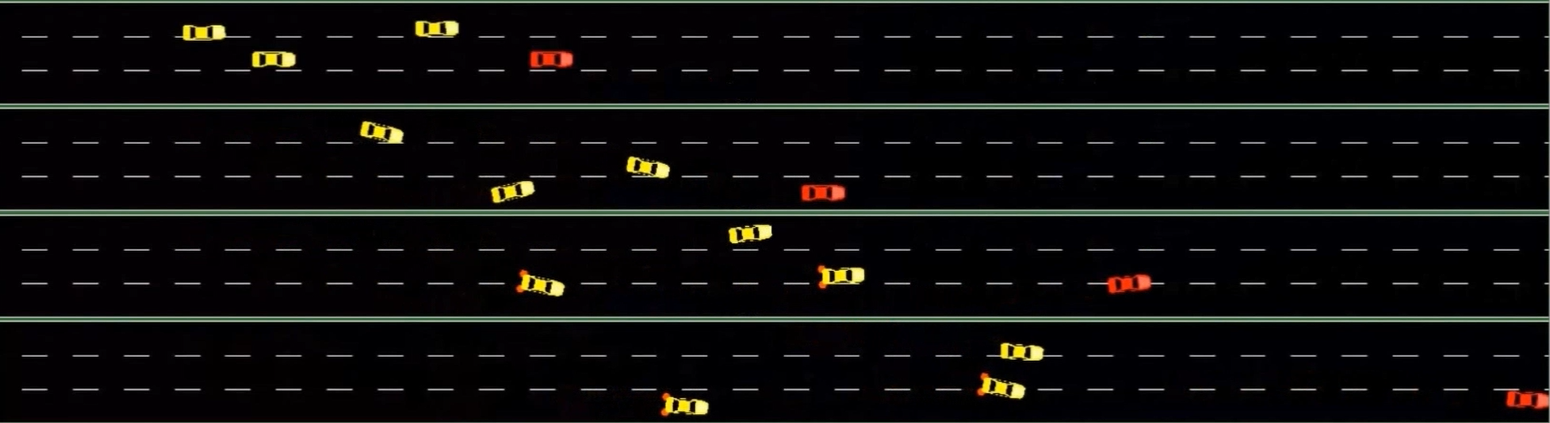}
        \end{subfigure}
    \end{minipage}
    \caption{\textit{Matrix Experiment}: A result in the matrix experiment with fixed \textit{predictor iteration} $=5$. According to the predicted labels of the fixed predictor, weighted sampling is performed to obtain the same test scenario. In the \textbf{top} image, \textit{AV iteration} $=1$, and the AV has experienced an accident; In the \textbf{bottom} image, \textit{AV iteration} $=10$, and the AV has successfully passed the test.}
    \label{fig:matrix_fix_predictor}
\end{figure*}
\begin{figure*}[h]
    \centering
    \begin{minipage}[H]{0.5\linewidth}
        \begin{subfigure}
            \centering
            \setlength{\abovecaptionskip}{0mm}
            \includegraphics[width=0.98\linewidth]{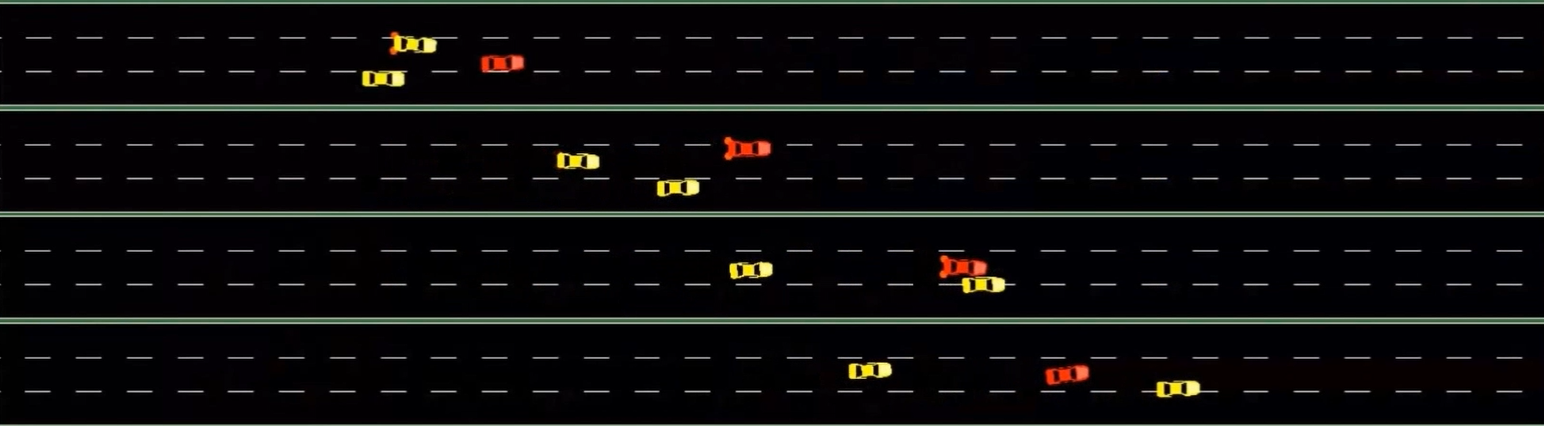}
        \end{subfigure}
    \end{minipage}
    \hfill
    \begin{minipage}[H]{0.5\linewidth}
        \begin{subfigure}
            \centering
            \setlength{\abovecaptionskip}{0mm}
            \includegraphics[width=0.98\linewidth]{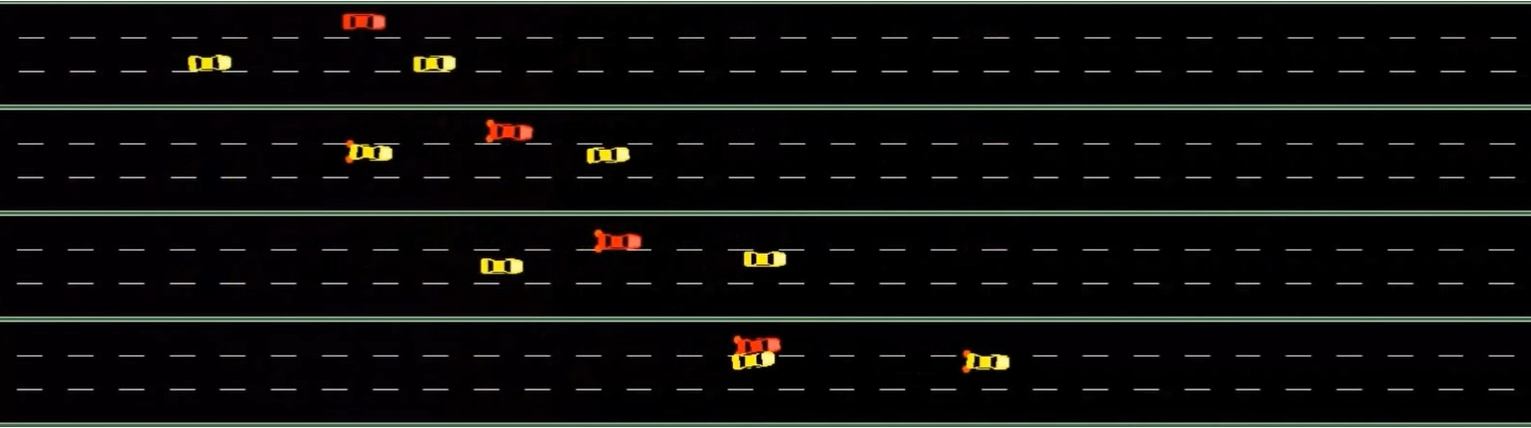}
        \end{subfigure}
    \end{minipage}
    \caption{\textit{Matrix Experiment}: A result in the matrix experiment with fixed \textit{AV iteration} $=5$. In the \textbf{top} image, \textit{predictor iteration} $=1$, and the selected scenario is relatively simple, where AV successfully passed the test; In the \textbf{bottom} image, \textit{predictor iteration} $=10$, the selected scenario is quite difficult, where the AV experienced an accident.}
    \label{fig:matrix_fix_AV}
\end{figure*}

\begin{figure*}[t]
    \centering
    \begin{minipage}[H]{0.5\linewidth}
        \begin{subfigure}
            \centering
            \setlength{\abovecaptionskip}{0mm}
            \includegraphics[width=0.98\linewidth]{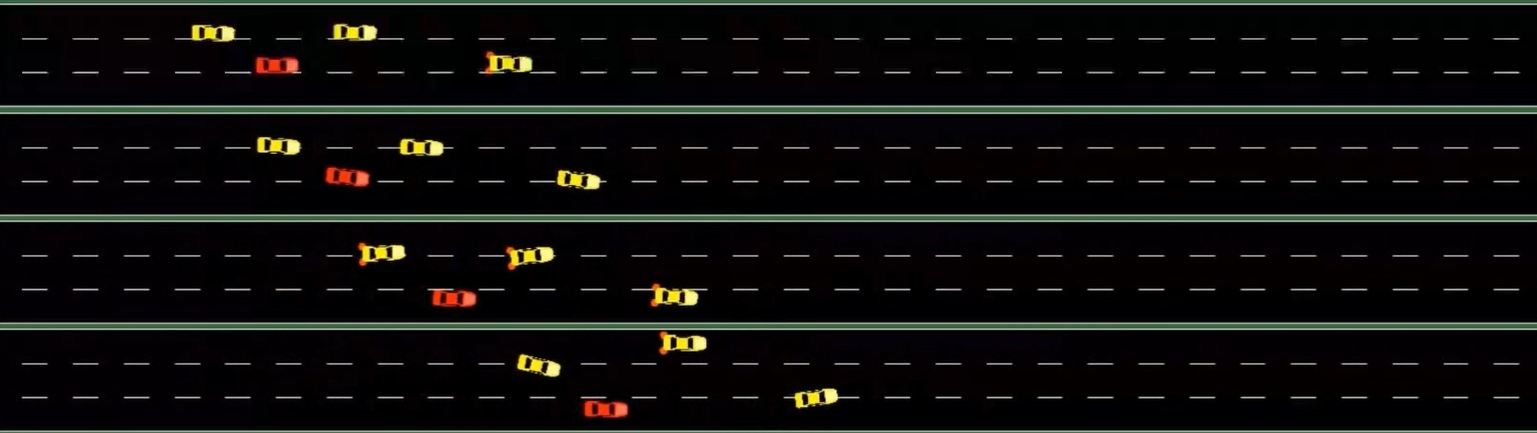}
        \end{subfigure}
    \end{minipage}
    \hfill
    \begin{minipage}[H]{0.5\linewidth}
        \begin{subfigure}
            \centering
            \setlength{\abovecaptionskip}{0mm}
            \includegraphics[width=0.98\linewidth]{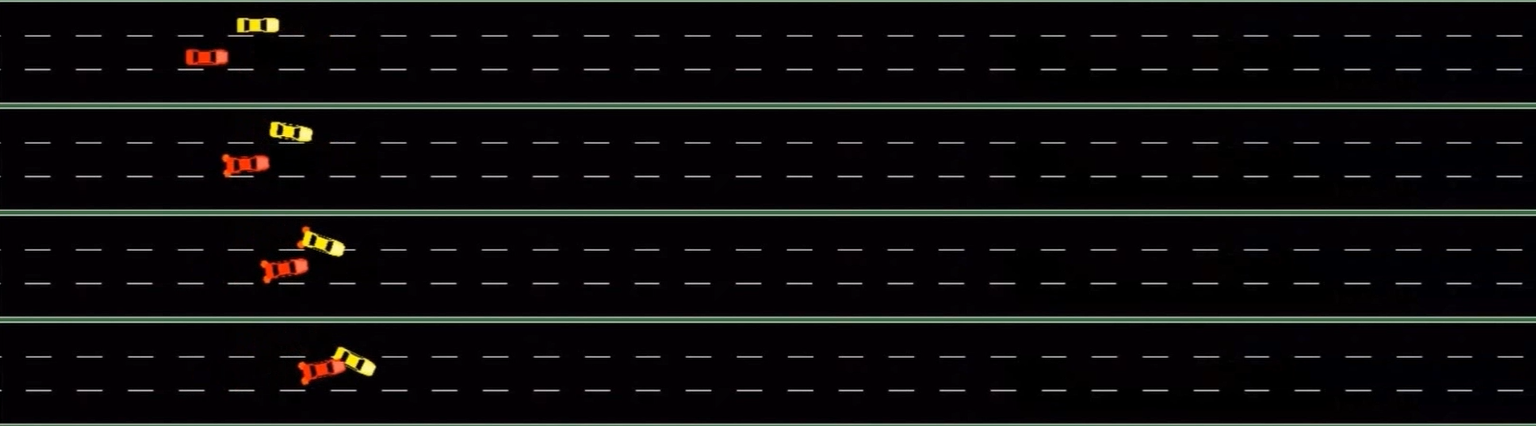}
        \end{subfigure}
    \end{minipage}
    \caption{\textit{Analyses on Individualization}: An individualized scenario selection result in Section~\ref{4-4}. On the \textbf{top} is one scenario selected for normal AV models, and on the \textbf{bottom} is one scenario selected for AV models with left front perception defects. It can be observed that in the scenario on the right, the BV in left front of the AV is significantly closer and more aggressive.}
    \vspace{155mm}
    \label{fig:individualization_vis}
\end{figure*}

\end{document}